\pdfoutput=1

\documentclass[11pt]{article}

\usepackage{acl}

\usepackage{times}
\usepackage{latexsym}
\usepackage[T1]{fontenc}
\usepackage[utf8]{inputenc}
\usepackage{microtype}
\usepackage{inconsolata}
\usepackage{graphicx}

\usepackage[ruled]{algorithm2e}
\usepackage{soul}
\usepackage{booktabs}
\usepackage{amsmath,caption}
\usepackage{colortbl}
\usepackage{arydshln}
\usepackage{multirow}
\usepackage{multicol}
\usepackage{hyperref}

\usepackage{xspace}
\usepackage{adjustbox}
\usepackage{subcaption}
\usepackage{longtable}
\usepackage{makecell}
\usepackage{bm}

\usepackage{listings}
\newcommand{\approach}{GPT-4V\xspace}

\title{Vision Language Models for Spreadsheet Understanding: \\
Challenges and Opportunities}

\author{
Shiyu Xia\thanks{\ \ Equal contribution.}\thanks{\ \ Work during internship at Microsoft.}, 
Junyu Xiong\footnotemark[1]\footnotemark[2], 
Haoyu Dong\thanks{\ \ Corresponding author.}, 
Jianbo Zhao\textsuperscript{\dag}, 
Yuzhang Tian\textsuperscript{\dag}, 
\\
\textbf{Mengyu Zhou,
Yeye He,
Shi Han,
Dongmei Zhang}
\\
Microsoft Corporation
}

\begin{document}
\maketitle
\begin{abstract}

This paper explores capabilities of Vision Language Models on spreadsheet comprehension. We propose three self-supervised challenges with corresponding evaluation metrics to comprehensively evaluate VLMs on Optical Character Recognition (OCR), spatial perception, and visual format recognition. Additionally, we utilize the spreadsheet table detection task to assess the overall performance of VLMs by integrating these challenges. To probe VLMs more finely, we propose three spreadsheet-to-image settings: column width adjustment, style change, and address augmentation. 

We propose variants of prompts to address the above tasks in different settings. Notably, to leverage the strengths of VLMs in understanding text rather than two-dimensional positioning, we propose to decode cell values on the four boundaries of the table in spreadsheet boundary detection. Our findings reveal that VLMs demonstrate promising OCR capabilities but produce unsatisfactory results due to cell omission and misalignment, and they notably exhibit insufficient spatial and format recognition skills, motivating future work to enhance VLMs' spreadsheet data comprehension capabilities using our methods to generate extensive spreadsheet-image pairs in various settings.

\end{abstract}

\begin{figure*}[h]
  \centering
  \includegraphics[width=\linewidth]{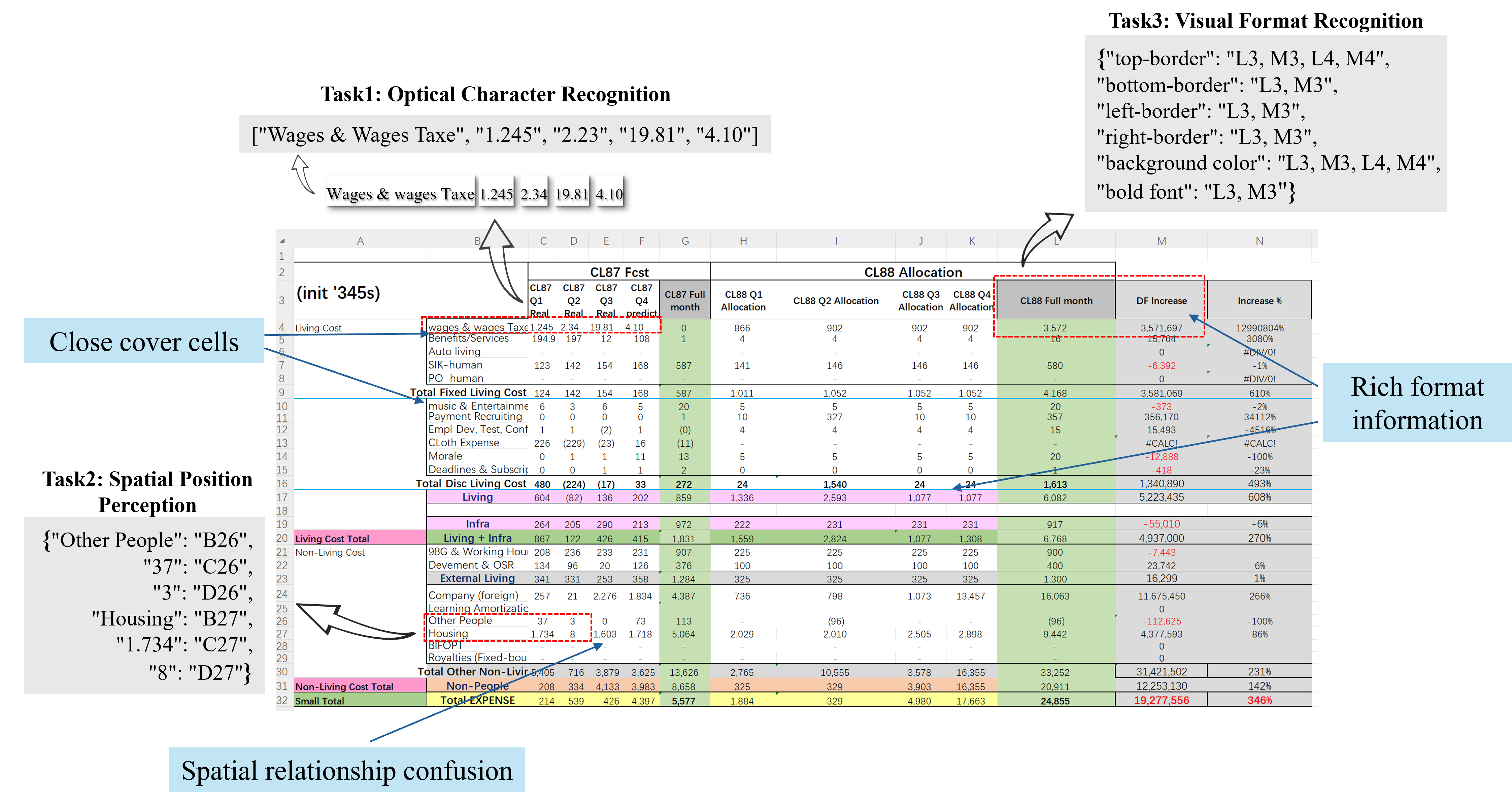}
  \caption{A sample spreadsheet showing various challenging points in spreadsheet understanding task.}
  \label{fig:dring case}
\end{figure*}

\section{Introduction}
Spreadsheets are widely-used for data management and analysis~\cite{birch2018future,wu2023huss}. However, they are designed to be "human-friendly, not "machine-friendly"~\ref{fig:dring case}. Cells are arranged on the grid and illustrated by various visual formats like borders, colors, and bold fonts. Unlike machines, humans naturally leverage these visual cues to understand the layouts and structures of spreadsheets, such as the location of the table (e.g., "A2:N32") using borders, the headers (e.g., "A2:N3") using bold fonts, and aggregated rows and columns (e.g., rows 17, 19, and 20) using fill colors.

While LLMs have shown promising performance in serializing spreadsheets as text sequences~\cite{chen2024sheetagent,li2024sheetcopilot}, representing spreadsheets in this manner loses critical visual signals. With the recent surge in Vision Language Models (VLMs)~\cite{laurenccon2024matters}, we propose studying the capability of language models to leverage visual signals for spreadsheet understanding.
Fortunately, a spreadsheet can be straightforwardly processed using third-party tools like Interop and converted into an image. This motivates us to construct spreadsheet-image pairwise data for self-supervised tasks. To this end, we propose three self-supervised tasks to comprehensively examine critical abilities of VLMs separately: Optical Character Recognition (OCR) of cells, two-dimensional spatial position perception, and visual format recognition.
Finally, we use spreadsheet table detection~\cite{dong2019tablesense}, a fundamental and enabling task in Microsoft Excel and Google Sheets, to jointly examine the effectiveness of VLMs, as this task combines the challenges of all three self-supervised tasks.

Specifically, as shown in Figure.~\ref{fig:dring case}, spreadsheet images present the following challenges: 1) The rows and columns are very compact, even overlapping, which makes the OCR task difficult. Specifically, VLMs sometimes struggle to split multiple cells and mistakenly treat them as a single cell. 2) The absence of explicit cell addresses and clear boundaries between rows and columns makes it difficult to perceive spatial locations. 3) Spreadsheets often contain a variety of formats, making it hard to recognize all formats precisely at the pixel level. To address these issues, we propose three different spreadsheet-to-image settings to probe the VLMs' performance: column width adjustment, style change, and address augmentation respectively, as shown in Figure.~\ref{fig:reconstructed method}.

We explore variants of prompts to address the above tasks in different settings. Notably, to leverage the strengths of VLMs in understanding text rather than two-dimensional positioning, we propose to decode cell values on the four boundaries of the table rather than decoding regions like "A2:E5" directly in the task of spreadsheet boundary detection.  By analyzing the experiment results, we draw the following conclusions: Firstly, VLMs possess strong OCR capabilities, yet they often encounter issues of cell omission and prediction misalignment when dealing with spreadsheet images. Secondly, VLMs lack robust spatial perception in images because they need to infer the number of rows and columns implicitly on a large two-dimensional cell grid rather than reading it directly. It is highly noteworthy that their performance on recognizing visual formats on a cell grid is far from satisfactory; they are far from human-level in comprehending spreadsheet formats. Lastly, in the task of spreadsheet table detection, VLMs do not perform as well as the existing CNN-based TableSense~\cite{dong2019tablesense}, which is well-trained using a human-labeled dataset, indicating that there is still a long way to go in understanding spreadsheet images for VLMs.

\section{Related Work}

\subsection{Table Representation}
The advent of Large Language Models (LLMs) has significantly spotlighted the task of processing structured data~\cite{jiang2023structgpt, tang2023struc, guo2023gpt4graph,dong2022table,dong2024large}, particularly tabular data. In the quest to effectively communicate tabular data to LLMs, researchers have devised numerous formats, including HTML, JSON, Markdown, and XML, to represent such data. Studies by Sui et al.~\cite{sui2023gpt4table} and Singha et al.~\cite{singha2023tabular} have underscored the efficacy of using Markdown and HTML for tabular data representation. However, these methods do not apply to spreadsheets since they have a single table assumption with an explicit region. Moreover, they do not leverage visual formats. \cite{deng2024tables} explored the usage of LLMs to evaluate representations of tables in image form, and Singh et al.~\cite{singh2023assessing} examined the capability of GPT-4 with vision(\approach)~\cite{achiam2023gpt} on structured data, but they also focus on table-based input but not spreadsheet input that can include multiple tables and scattered notes. In contrast, there's a growing interest in exploring the vision perspective of spreadsheets to leverage the visual cues and take the whole spreadsheet rather than a single table as input. For instance, Dong et al.~\cite{dong2019tablesense} uses CNN to capture spatial layouts of spreadsheet.~\cite{wang2021tuta} uses transformer-based encoders to learn embeddings of cell values and formats in spreadsheets. However, our research diverges by focusing on exploring LLMs' ability to understand spreadsheet images. \cite{huang2023improving} proposed to model table boundaries as language sequences and use sequence decoder for table recognition. 

\subsection{Table-Related Tasks}
Previous research has extensively explored tasks related to tables, encompassing table QA, table fact-checking, table-to-text, table manipulation, and table interpretation, etc~\cite{pasupat2015compositional,novikova2017e2e,chen2020tabfact,sui2023tap4llm, li2023table, zhang2023tablellama}. However, many of these tasks primarily revolve around understanding tables at the textual level. In reality, tables are often embedded within documents, images, and web pages, necessitating the exploration of related tasks such as table header detection, table structure recognition, and table recognition.

In recent studies, Fang et al.~\cite{fang2012table} identified tables within PDF documents using existing table extraction tools and employed machine learning algorithms to construct classifiers for identifying and categorizing table headers. Nassar et al.~\cite{nassartableformer} introduced a novel table unit object detection decoder based on Transformer architecture to comprehend table structures. Ly et al.~\cite{ly2023end} decomposed the table recognition task into two subtasks: table structure recognition and cell content recognition. They proposed an end-to-end multi-task learning model to address these subtasks.

However, our current study focuses more on the understanding of spreadsheet images by VLMs. This involves investigating the OCR capabilities of VLMs, their aptitude in capturing formatting information, their perception of spatial positioning, and their efficacy in detecting tables from spreadsheets~\cite{dong2019tablesense}.

\section{Preliminary}

\subsection{Probing tasks}
We design the following three probing tasks to evaluate the performance of VLMs on spreadsheet understanding.

\textbf{Optical Character Recognition (OCR):}
A spreadsheet is a two-dimensional cell grid that differs from plain text. In OCR tasks for text, the output simply sequences the characters. However, OCR for spreadsheets not only involves recognizing characters but also requires organizing them in units of distinct cells as shown in Task1 of Figure.~\ref{fig:dring case}.

\textbf{Understanding spatial position:}
The ability of VLMs to perceive the spatial position of images has been a long-standing challenge. Unlike ordinary images, spreadsheet images employ a precise two-dimensional coordinate system, where misalignment of rows and columns severely disrupts the understanding of information. Each cell's address corresponds to exact row and column coordinates, however, the images don't explicitly indicate the coordinate positions, so we define the top row in the image as the first row and the leftmost column as the first column. Consequently, the address of the cell located at the intersection of the first row and first column is defined as "1,1". Cell numbers increase from left to right and from top to bottom.
As shown in Task2 of Figure.~\ref{fig:dring case}, the address for "Other People" is "B26." But for spreadsheet images without given coordinate positions, it should be recognized as "26,2".

\textbf{Understanding visual format information:}
Spreadsheets contain rich formatting details that enhance comprehension and processing. If VLMs could "read" format information in images, it would perceive the images much like human do. Although spreadsheets contains a variety of format, we primarily focus on top border, bottom border, left border, right border, bold font, and fill color as shown in Task3 of Figure.~\ref{fig:dring case}. 

\subsection{Spreadsheet Table Detection Task}
Spreadsheet table detection ~\cite{dong2019tablesense}, involves identifying all tables within a given spreadsheet and determining their respective ranges. The spreadsheet will feature a visually rich design containing several tables scattered throughout, each potentially featuring a unique structure. Variability in the layout and structure of multiple tables contains rich visual information greatly complicating the task by obscuring table boundaries. Spreadsheet table detection is a horizontal and enabling task benefiting various intelligent features in spreadsheet softwares. Therefore, We employ this critical task in our work to assess the extent to how visual information influences the ability of VLMs to comprehend spreadsheets.

\section{Methodology}

\subsection{Dataset Construction}

In order to study the spreadsheet understanding capabilities of VLMs such as \approach and Gemini~\cite{team2023gemini}, we convert the spreadsheet dataset~\cite{dong2019tablesense} into images using the Microsoft Office Interop Excel library~\footnote{https://github.com/microsoft/Windows-Packaging-Samples/tree/master/OfficeInterop/Excel.Interop} without any human labeling efforts. Then we can simply reverse the dataset to get image-spreadsheet pairs.

\begin{figure}
  \centering
  \includegraphics[width=\linewidth]{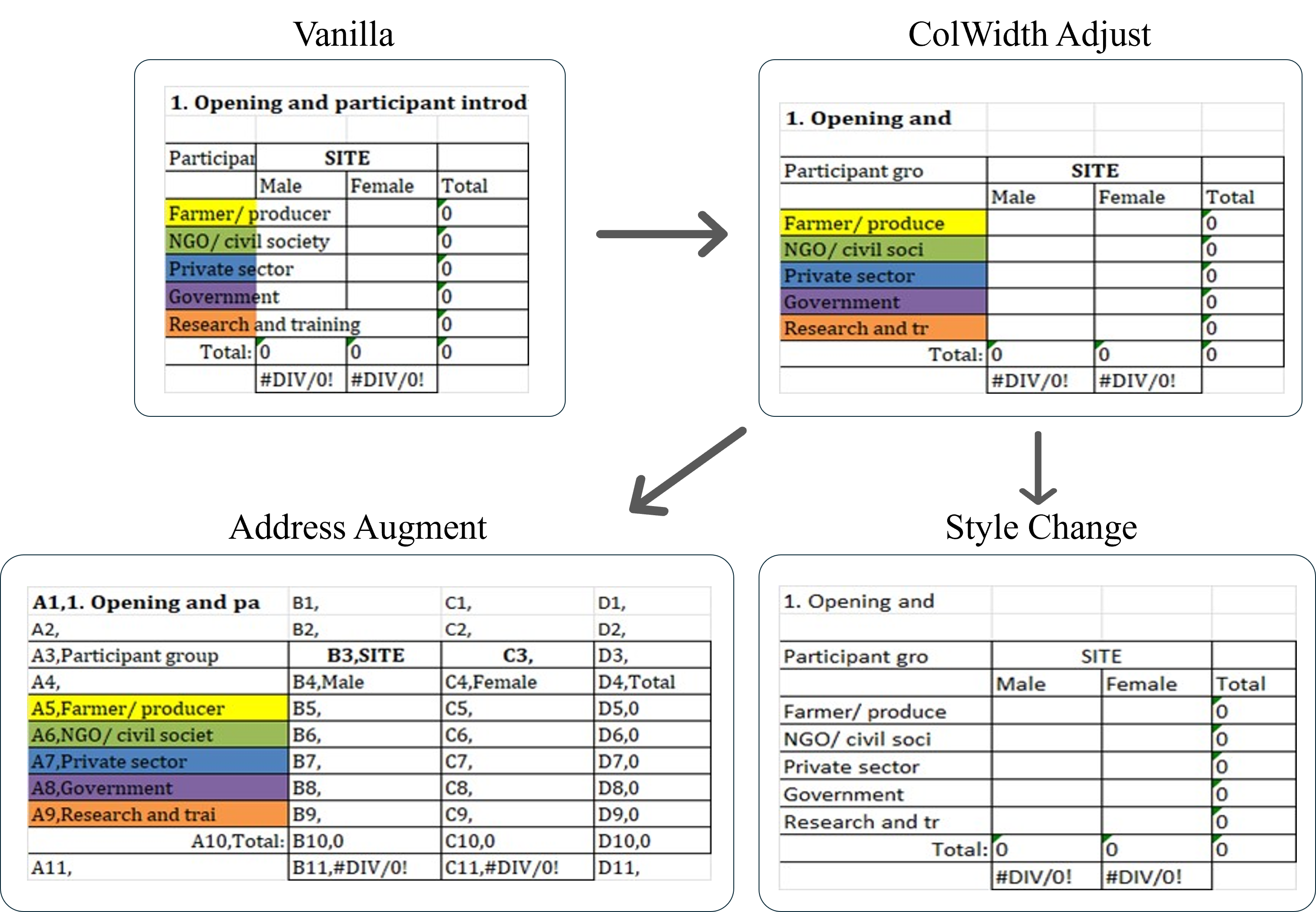}
  \caption{Illustration of spreadsheet-to-image settings.}
  \label{fig:reconstructed method}
\end{figure}

Next, to probe the differences for VLMs to understand image spreadsheets under various image settings, we propose three processing methods on the input spreadsheet shown as Figure.~\ref{fig:reconstructed method}. They are column width adjustment, style change, and address augment, respectively.

\textbf{Column Width Adjustment:} 
Since the column width effect the maximum number of characters displayed in each cell, if the column width is too small, the content between multiple cells will be very compact, making it difficult for the model (or even humans) to distinguish it. If the column width is too large, space will be wasted. 
Therefore, we come up with a setting that adjusts the column width based on the text length, but if the text length is too long, we limit it to the first 15 characters.

\textbf{Style Change:}
Spreadsheet style attributes mainly include background color and various font properties such as bold, italic, fill color, and size. These styling elements serve specific functions, for instance, background color often groups similar data, while font color and bolding emphasize important details. These styles provide distinct visual cues within the spreadsheet. To minimize the influence of these stylistic elements on the understanding of VLMs, it's necessary to standardize these attributes: removing background colors and bold formatting from each cell, setting font color to black, and using a consistent font type and size, such as Calibri at 12pt, etc.

\textbf{Address Augment:}
In spreadsheets, cell contents typically serve the sole purpose of storing data. However, a comprehensive understanding of the spreadsheet requires grasping the spatial relationships and format correspondences between cells. Existing VLMs may struggle to robustly capture these precise spatial relationships. To address this, we propose a new setting that incorporates cell address information alongside the cell content. That is, we explicitly concatenate the cell address (e.g., "A1") with its value (e.g., "day"), using a comma to separate them. This results in a fashion like "A1, day." 

\subsection{Optical Character Recognition}

\begin{figure}
    \centering
   
    \includegraphics[width=\linewidth]{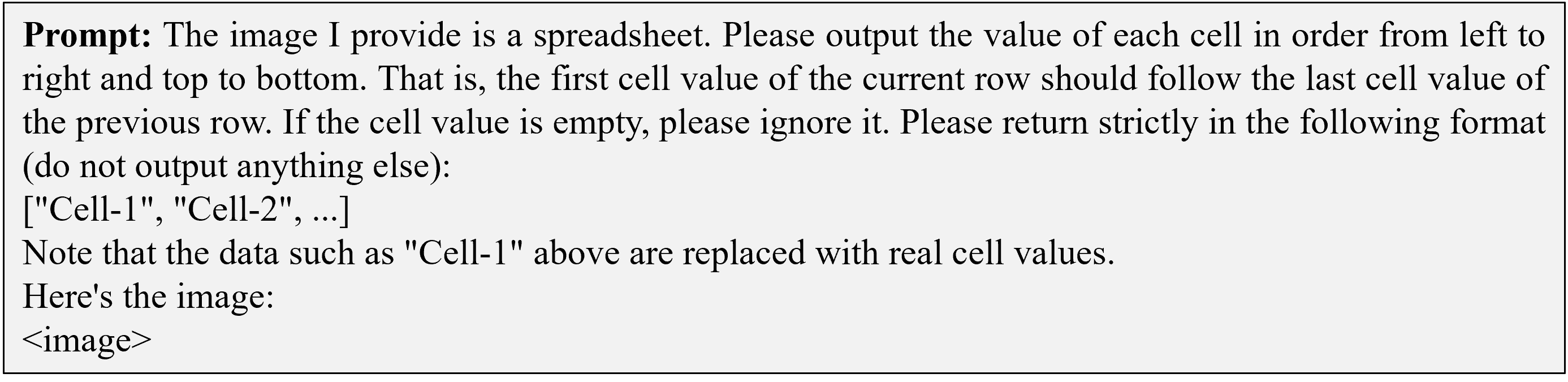}
    
    \caption{The prompt of OCR task.}
    \label{fig:ocr_prompt}
\end{figure}

We instruct the VLM to sequentially decode the text of each cell in the spreadsheet image, moving from top to bottom and left to right, while omitting cells that contain null values.  Figure.~\ref{fig:ocr_prompt} provides a prompt example.

\begin{figure}[h]
  \centering
  \includegraphics[width=1.0\linewidth]{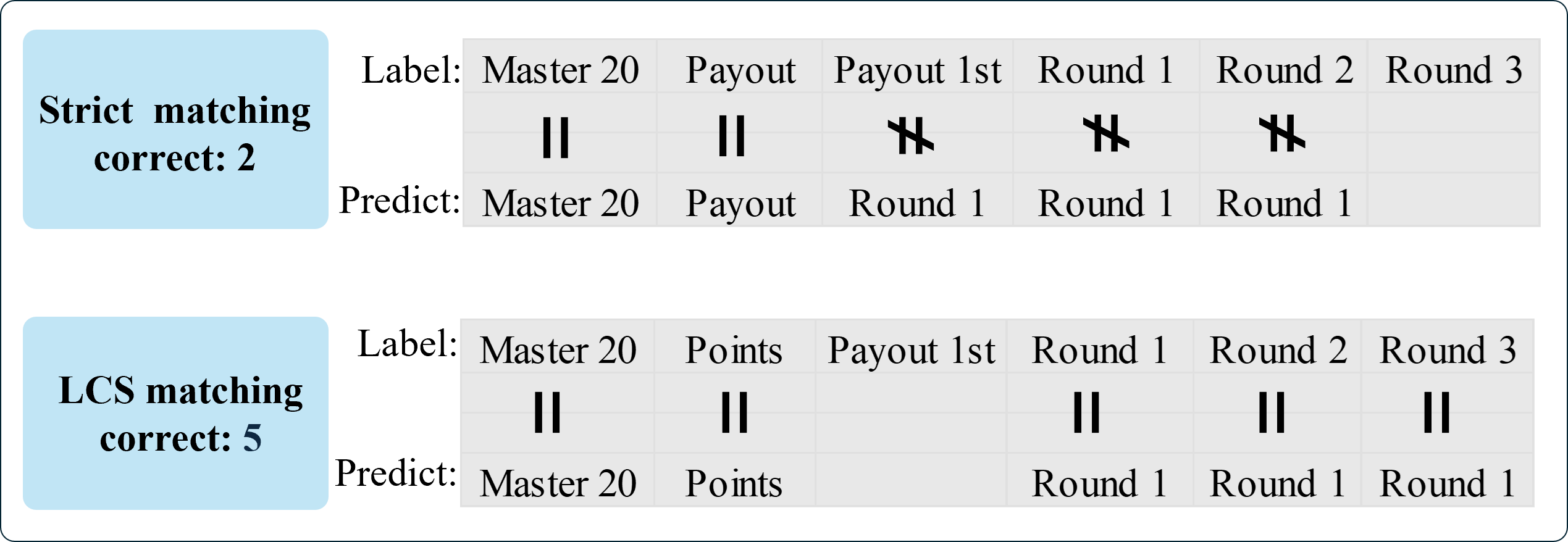}
  \caption{The difference between LCS matching and Strict matching.}
  \label{fig:strict_LCS}
\end{figure}

\textbf{Evaluation Method:}
We adopt two kinds of evaluation method, Strict and longest common substring (LCS). As shown in Figure.~\ref{fig:strict_LCS}, the LCS argorithm is uesd to find the longest common subsequence between the predicted sequence and the ground truth sequence. It helps to effectively alleviate the problem of poor performance caused by missing some cells in the output and can test the OCR ability of the VLMs to the greatest extent. 

\subsection{Spatial Position Perception}

We prompt the VLMs to recognize the spatial positions of specified cells ensuring that each cell value and its address correspond uniquely. Figure.~\ref{fig:Spatial Position Perception_prompt} provides a prompt for vanilla experiment, other prompts see Appendix~\ref{appendix: A}.

Specifically, we input a spreadsheet image along with a list of randomly shuffled cell values into the VLMs. Then, we prompt the VLMs to output the address corresponding to each value. It is important to note that for the vanilla, colwidth adjust, and style change experiments, the input image does not contain cell addresses. Therefore, the addresses output by the VLMs should be composed of the row and column indices of the cell, in the form "2,3". In contrast, the address augment experiment outputs addresses in the form "C2".

\begin{figure}
    \centering
    \includegraphics[width=\linewidth]{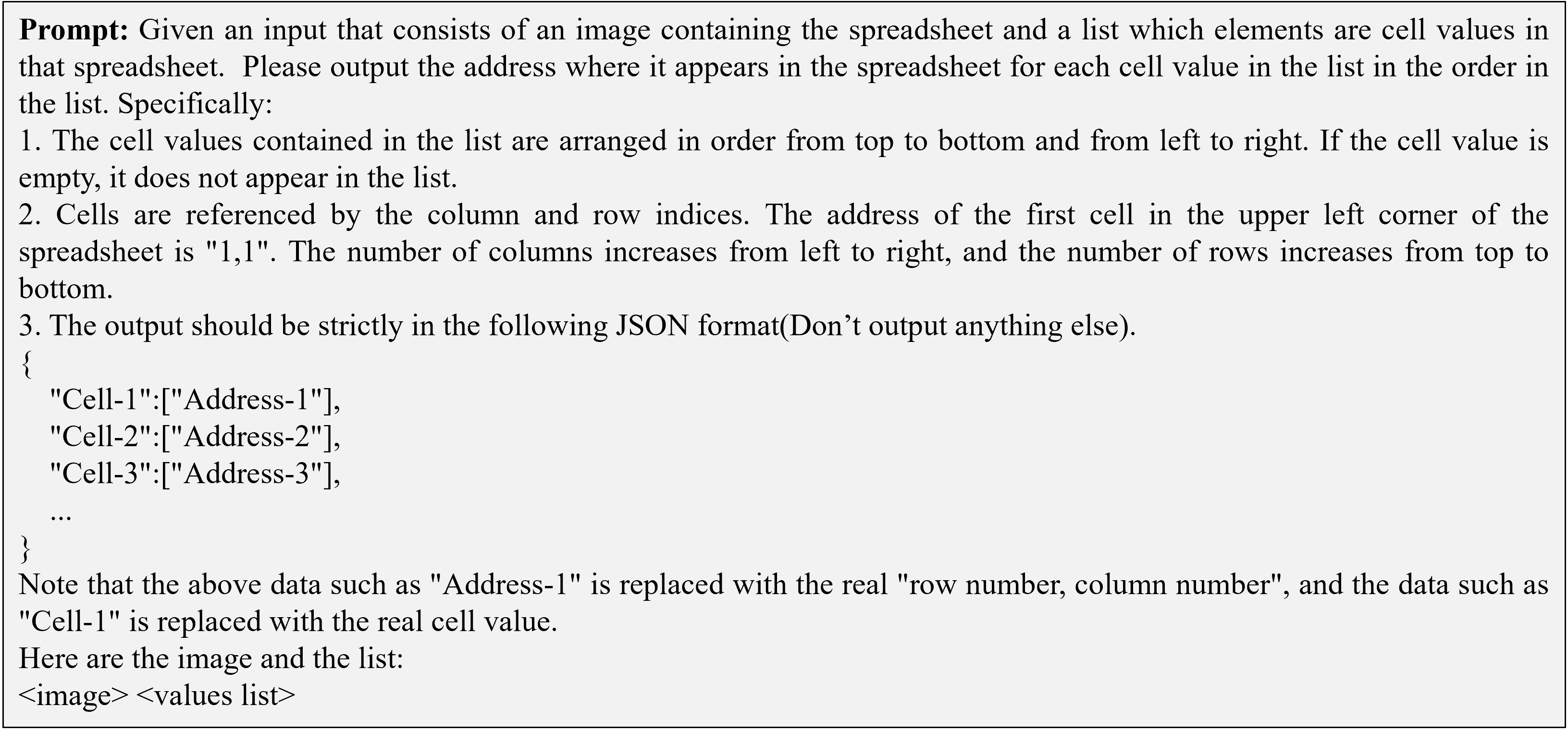}
    \caption{The prompt for vanilla experiment of spatial position perception task.}
    \label{fig:Spatial Position Perception_prompt}
\end{figure}

\subsection{Visual Format Recognition}
We have defined six specific cell formats: top border, bottom border, left border, right border, bold font, and fill color. For each format, we instruct the VLMs to identify and output the addresses of all cells that exhibit the specified format. Figure.~\ref{fig:Visual Format Recognition_prompt} provides a prompt for vanilla experiment, other prompts see Appendix~\ref{appendix: A}.

This experiment is similar to the spatial position perception experiment. The addresses output by the vanilla and colwidth adjust experiments should be composed of the row and column indices of the cell in the form "1,2", while the address augment experiment outputs addresses in the form "B1".

\begin{figure}
    \centering
    \includegraphics[width=\linewidth]{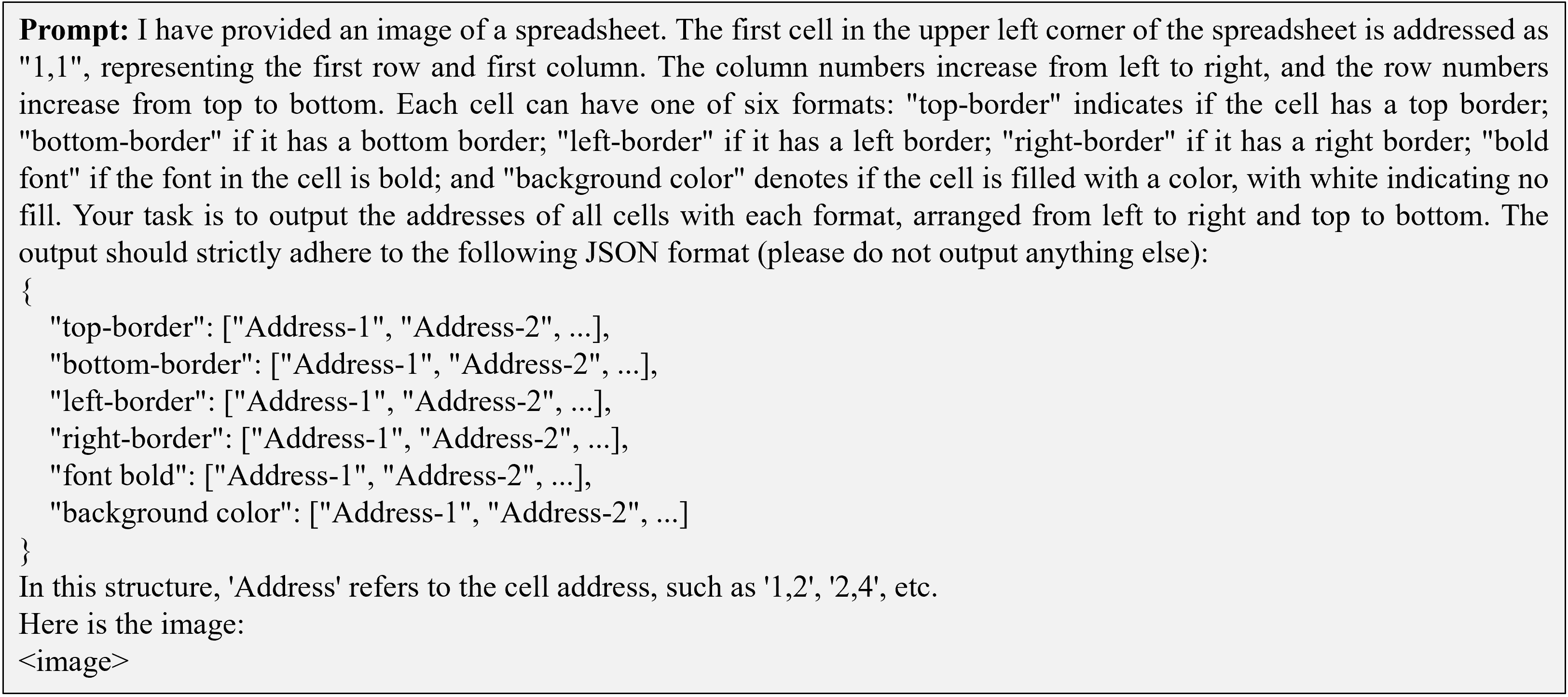}
    \caption{The prompt for vanilla experiment of visual format recognition task.}
    \label{fig:Visual Format Recognition_prompt}
\end{figure}

\subsection{Spreadsheet Table Detection}
We instruct VLMs to detect all table ranges from spreadsheet images.  Figure.~\ref{fig:TD_prompt} provides a prompt for vanilla experiment, other prompts see Appendix~\ref{appendix: A}.

By convention, contiguous cell ranges are represented by the addresses of the upper left and lower right cells, separated by ":", and cells are referenced by their column and row indices, e.g., "A4:D120". However, when presenting a spreadsheet as an image input to the VLMs, the image may lack the ability to deduce the cell addresses. To address this challenge, we propose a novel approach where the VLMs directly decode the contents of the four boundaries of the table. Subsequently, these decoded contents are mapped to a conventional addresses using our proposed method as introduced in the follow paragraph. 

Specifically, except for the address augment experiment, which can directly output a range in the form "A4:D120," the other experiments output the result by decoding the four boundaries.

\textbf{Mapping Algorithm:}
Consider a spreadsheet $S$ comprising $m$ rows and $n$ columns, where each cell is represented by $c_{i,j}$, with $i$ and $j$ denoting its row and column index, respectively, within the spreadsheet. 

\begin{equation}
 S = \left[ \begin{array}{cccc}
c_{1,1} & c_{1,2} & \cdots & c_{1,n} \\
c_{2,1} & c_{2,2} & \cdots & c_{2,n} \\
\vdots  & \vdots  & \ddots & \vdots  \\
c_{m,1} & c_{m,2} & \cdots & c_{m,n} \\
\end{array} \right]
\end{equation}
By allowing the model to decode the four boundaries of all tables, we obtain the model's prediction result denoted as $Predict=[T_1, T_2, ...]$. 
Among them, $T_i$ means the predicted four boundaries of the $i$-th table, that is,
        
\begin{equation}
\begin{aligned}
    T_i = \left\{
        \begin{array}{l}
            B_t :  [c_1, c_2, \dots, c_t], \\
            B_b :  [c_1, c_2, \dots, c_b], \\
            B_l :  [c_1, c_2, \dots, c_l], \\
            B_r :  [c_1, c_2, \dots, c_r]
        \end{array}
    \right\}
\end{aligned}
\end{equation}

Among them, $B_t$, $B_b$, $B_l$ and $B_r$ represent the contents of $top\_border$, $bottom\_borde$r, $left\_border$ and $right\_border$ respectively. 
For top\_border and bottom\_border, we can map it to the most likely row index in the spreadsheet through algorithm~\ref{alg:four2address}. For left\_border and right\_border, we only need to transpose them and do the same. Finally, after we obtain the row/column index corresponding to each predicted border, we process it into a region such as “A1:D9”.

\RestyleAlgo{ruled}
\SetKwComment{Comment}{/* }{ */}
\begin{algorithm}[hbt!]
\caption{Map the content of a specific row to the corresponding row index.}
\label{alg:four2address}
\SetKwInOut{Input}{Input}
\SetKwInOut{Output}{Output}
\Input{The border content $B$ predicted by the model and the contents $S$ of the spreadsheet. }
Initialize the origin confidence $Conf$ to $0.8$.\;
Initial the result index $res$ to $-1$.\;
\For{$i = 1$ \KwTo $|S|$}{
    \If{$|S[i]| \geq |B|$}{
        $tList \leftarrow S[i]$\;
        $sList \leftarrow B$\;
    }
    \Else{
        $tList \leftarrow B$\;
        $sList \leftarrow S[i]$\;
    }
    $sCnt \leftarrow |sList|$\;
    $tCnt \leftarrow |tList|$\;
    \For{$j = 1$ \KwTo $tCnt$}{
        \If{$j + shortCnt > tCnt$}{
            Break\;
        }
        $cConf \leftarrow \frac{\sum_{k=1}^{sCnt} (sList[k] == tList[k])}{\text{sCnt}}$\;
        \If{$cConf \geq Conf$}{
            $res \leftarrow i$\;
            $Conf \leftarrow cConf$\;
        }
    }
}
\Output{the result index $res$.}
\end{algorithm}

\begin{figure}
    \centering
    \includegraphics[width=\linewidth]{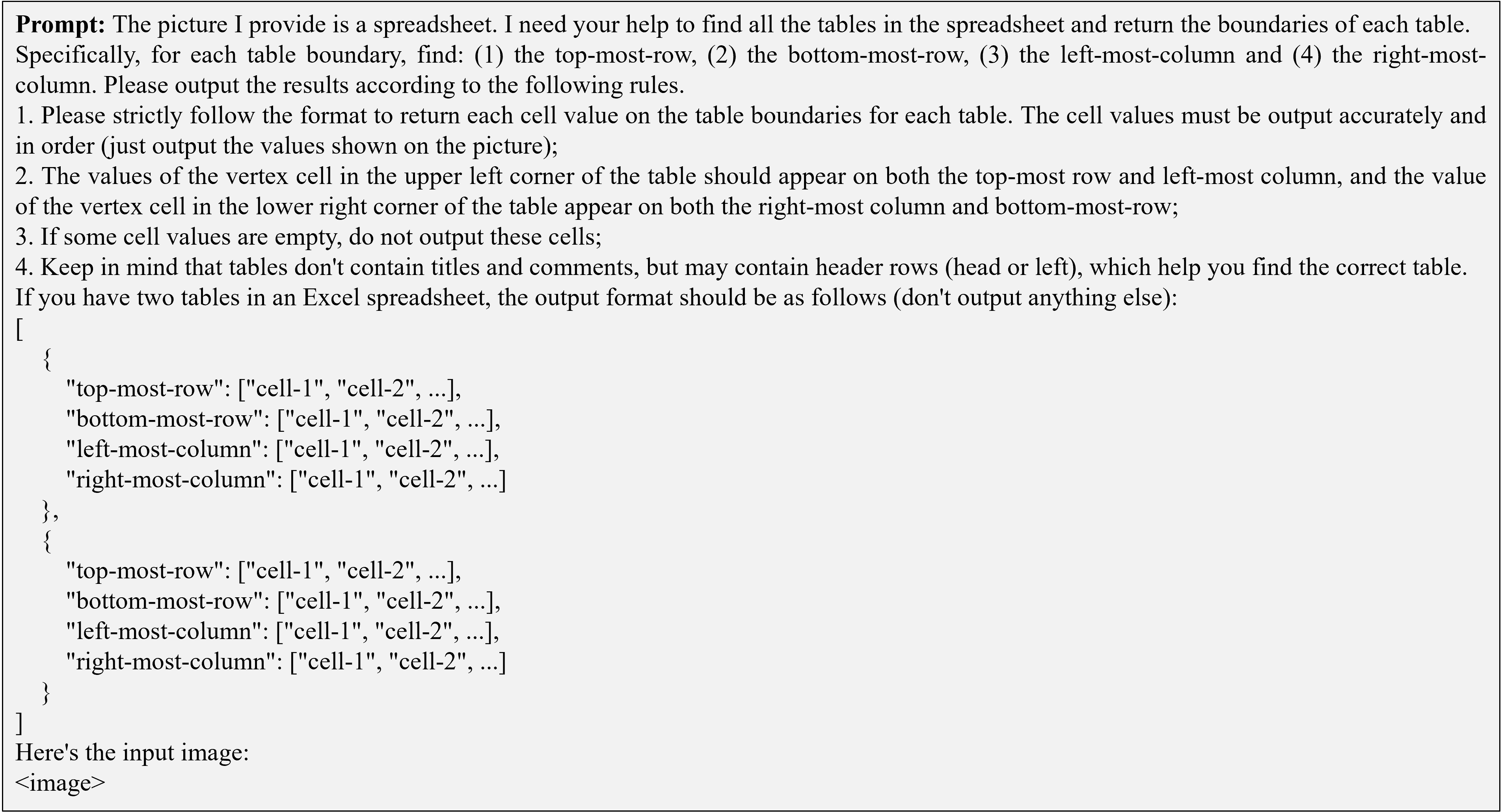}
    \caption{Zero-shot prompt for vanilla images decoding four boundaries on spreadsheet table detection task.}
    \label{fig:TD_prompt}
\end{figure}

\section{Experiment Setting}

We conducted experiments using \approach (\textit{2024-02-15 preview})~\cite{achiam2023gpt} and Gemini (\textit{1.5-pro lateset until 2024-05-16})~\cite{team2023gemini}. To ensure consistent experimental parameters, we set the generation temperature for both \approach and Gemini-pro to 0.7, top\_p to 0.95, and max\_output\_token to 4096.

Due to \approach's input image restrictions—specifically, that the image file size must be less than 4MB and the resolution must be between $50 \times 50$ and $10,000 \times 10,000$ pixels—we filtered 76 images from the test dataset of TableSense~\cite{dong2019tablesense} to meet these criteria. For each experiment, we evaluate the models using precision, recall, and F1, repeating each experiment three times and taking the average results.

\section{Experiment Result}

\begin{table*}[ht]
\centering
\begin{tabular}{llcccccc}
\toprule
\multirow{2}{*}{\%} & & \multicolumn{3}{c}{Strict match} & \multicolumn{3}{c}{LCS match} \\
\cmidrule(lr){3-5}
\cmidrule(lr){6-8}
& & Precision & Recall & F1 & Precision & Recall & F1 \\

\midrule

\multirow{3}{*}{\textbf{\approach}} 
&Vanilla & 14.78 & 12.68 & 13.65 & 74.32 & 63.74 & 68.63 \\
&ColWidth Adjust & 17.87 & 15.24 & 16.96 & 83.87 & 75.74 & 79.59 \\
&Style Change & 27.44 & 25.54 & 26.46 & 82.40 & 76.69 & 79.44 \\
\midrule
\multirow{3}{*}{\textbf{Gemini-pro}} 
&Vanilla & 9.26 & 8.08 & 8.63 & 80.61 & 70.39 & 75.14 \\
&ColWidth Adjust & 16.13 & 13.03 & 14.42 & 89.03 & 71.94 & 79.58 \\
&Style Change & 16.40 & 13.74 & 14.95 & 89.80 & 75.20 & 81.85 \\
\bottomrule
\end{tabular}

\smallskip 
\caption{Precision, recall and F1 results of GPT-4V and Gemini-pro on OCR task. Among them, colwidth adjust is the processing operation of column width adjustment. }  
\label{table: OCR-result}
\end{table*}

\subsection{Performance of Optical Character Recognition}
Table.~\ref{table: OCR-result} presents the OCR task results of GPT-4V and Gemini-pro, calculated using both Strict and LCS matching methods. From the table, we can observe:

1) Both GPT-4V and Gemini-pro are generally capable of accurately recognizing the content in spreadsheet images. Specifically, the best performance of GPT-4V and Gemini-pro can reach F1 scores of 79.59\% and 81.85\%, respectively, demonstrating their strong ability to recognize content in a two-dimensional grid.

2) For both GPT-4V and Gemini-pro, the performance of LCS matching far exceeds that of Strict matching, indicating that they tend to miss some cells or predict misalignments during performing OCR task, causing almost all predictions to be incorrect from the first missed cell in Strict matching. Specifically, GPT-4V's F1 scores under LCS matching are higher than Strict matching by 54.98\%, 62.63\%, and 52.98\% and Gemini-pro's F1 scores under LCS matching are higher by 66.51\%, 65.16\%, and 66.9\% for the three different inputs, respectively.

3) Preprocessing spreadsheets by adjusting column width significantly enhances the OCR capabilities of VLMs on spreadsheet images, but further preprocessing with style change does not improve the OCR performance of VLMs. Specifically, adjusting column width can increase GPT-4V's F1 scores by 3.31\% and 10.96\% in the Strict match and LCS match methods, respectively, and increase Gemini-pro's F1 scores by 5.79\% and 4.44\% in the Strict match and LCS match methods, respectively.

4) Gemini-pro's OCR capability on spreadsheet images is slightly stronger than that of \approach. Specifically, in the vanilla and style change experiments, Gemini-pro's F1 scores are 6.51\% and 2.41\% higher than those of \approach, respectively.

Finally, we analyze the results of GPT4-V on a case in detail in Appendix~\ref{Appendix B: ocr}.

\begin{table}[h]
\centering
\scalebox{0.7}{
\begin{tabular}{llccc}
\toprule
\% & & Precision & Recall & F1 \\
\midrule

\multirow{4}{*}{\textbf{\approach}} 
& $\text{Vanilla}_{\text{\,number}}$ & 12.44 & 12.37 & 12.41 \\
& $\text{ColWidth Adjust}_{\text{\,number}}$ & 13.45 & 13.35 & 13.39 \\
& $\text{Style Change}_{\text{\,number}}$ & 12.16 & 12.14 & 12.15 \\
& $\text{Address Augment}_{\text{\,address}}$ & 48.87 & 49.09 & 48.97 \\
\midrule
\multirow{4}{*}{\textbf{Gemini-pro}} 
& $\text{Vanilla}_{\text{\,number}}$ & 16.72 & 18.00 & 17.33 \\
& $\text{ColWidth Adjust}_{\text{\,number}}$ & 14.29 & 15.40 & 14.82 \\
& $\text{Style Change}_{\text{\,number}}$ & 16.75 & 18.13 & 17.41 \\
& $\text{Address Augment}_{\text{\,address}}$ & 83.66 & 87.53 & 85.55 \\
\bottomrule
\end{tabular}
}

\smallskip 
\caption{Precision, recall and F1 results of GPT-4V and Gemini-pro on spatial position perception task. }  
\label{table: spatial position perception}
\end{table}

\begin{table}[h]
\centering
\scalebox{0.7}{
\begin{tabular}{llccc}
\toprule
\% & & Precision & Recall & F1 \\
\midrule
\multirow{2}{*}{\textbf{\approach}} 
& $\text{Vanilla}_{\text{\,number}}$ & 24.97 & 11.69 & 15.79 \\
& $\text{ColWidth Adjust}_{\text{\,number}}$ & 24.31 & 11.07 & 14.77\\

& $\text{Address Augment}_{\text{\,address}}$ & 28.88 & 13.28 & 17.83 \\
\midrule
\multirow{2}{*}{\textbf{Gemini-pro}} 
& $\text{Vanilla}_{\text{\,number}}$ & 35.27 & 13.28 & 17.53 \\
& $\text{ColWidth Adjust}_{\text{\,number}}$ & 35.09 & 12.69 & 16.93 \\
& $\text{Address Augment}_{\text{\,address}}$ & 41.93 & 16.78 & 22.19 \\
\bottomrule
\end{tabular}
}

\smallskip 
\caption{Precision, recall and F1 results of GPT-4V and Gemini-pro on visual format recognition task. }  
\label{table: format information understand}
\end{table}

\begin{table*}[h]
\centering
\scalebox{0.8}{
\begin{tabular}{clccccccccc}
\toprule
\multirow{2}{*}{\%}&  & \multicolumn{3}{c}{\textbf{Zero-Shot}} & \multicolumn{3}{c}{\textbf{One-Shot}} & \multicolumn{3}{c}{\textbf{Trained}}\\ 
\cmidrule(lr){3-5}
\cmidrule(lr){6-8}
\cmidrule(lr){9-11}
&   & Precision & Recall & F1 & Precision & Recall & F1 &  Precision & Recall & F1\\
\midrule

\multirow{3}{*}{\textbf{\approach}} & $\text{Vanilla}_{\text{\,four}}$ & 52.38 & 10.23 & 17.11 & 49.29 & 8.66 & 14.68  &-&-&-\\
& $\text{ColWidth Adjust}_{\text{\,four}}$ & 49.43 & 17.49 & 25.79 & 48.72 & 15.10 & 23.05 &-&-&-\\
& $\text{Address Augment}_{\text{\,range}}$ & 9.26 & 14.85 & 11.41 & 14.60 & 13.86 & 14.22  &-&-&-\\

\cmidrule(lr){1-11}

 \multirow{3}{*}{\textbf{Gemini-pro}} & $\text{Vanilla}_{\text{\,four}}$ & 25.96 & 18.40 & 21.53 & 35.82 & 6.98 & 11.67  &-&-&-\\
& $\text{ColWidth Adjust}_{\text{\,four}}$ & 26.66 & 22.40 & 24.33 & 26.93 & 7.03 & 11.15  &-&-&-\\
& $\text{Address Augment}_{\text{\,range}}$  & 9.08 & 19.94 & 12.47 & 7.62 & 15.55 & 10.00  &-&-&-\\
\cmidrule(lr){1-11}


\textbf{TableSense} & Text Input & - & - & - & - & - & -  &80.21 & 76.24 & 78.17\\
\bottomrule
\end{tabular}
}
\smallskip 
\caption{Precision, recall and F1 results of GPT-4V, Gemini-pro and TableSense~\cite{dong2019tablesense} on spreadsheet table detection task. }  
\label{table:table detection task}
\end{table*}

\subsection{Performance of Spatial Position Perception}
Table~\ref{table: spatial position perception} shows the results of GPT-4V and Gemini-pro in performing spatial position perception tasks. Analyzing the results in Table ~\ref{table: spatial position perception}, we first observe that GPT-4V and Gemini-pro perform poorly in the vanilla, colwidth adjust, and style change experiments. This underperformance is attributed to the three types of experiments demanding that the VLMs count the rows and columns in the spreadsheet. However, the boundaries of rows and columns in the spreadsheet are often unclear due to the lack of borders or the presence of line breaks that cause content overlap (e.g., "A5", "C3", etc. in Figure.~\ref{fig:dring case}).

Secondly, we noted that although preprocessing spreadsheets with address augment can significantly enhance the performance of both \approach and Gemini-pro, since address augment allows VLMs to fully utilize their OCR capabilities, GPT-4V does not achieve the same level of OCR performance as Gemini-pro. This suggests that GPT-4V may not understand the task prompts as thoroughly as Gemini-pro.

 In addition, we observe that in the four types of experiments, Gemini-pro outperform GPT-4V in F1 scores by 4.92\%, 1.43\%, 5.26\%, and 36.58\%, respectively, indicating that Gemini-pro has a stronger spatial position perception capability in spreadsheet image tasks.

 Finally, we analyze the results of GPT4-V on a case in detail in Appendix~\ref{Appendix B: spatial position perception}.

\subsection{Performance of Visual Format Recognition}
Table~\ref{table: format information understand} presents the results of GPT-4V and Gemini-pro in testing their ability to recognition the visual format information in spreadsheet images. The results indicate that the best F1 scores for GPT-4V and Gemini-pro across multiple experiments are only 17.83\% and 22.19\%, respectively. This demonstrates that their ability to comprehend format information in images is quite poor and that they cannot deeply understand images by combining format information as humans do. Therefore, this is an area where VLMs need improvement in the future. Additionally, in two types of experiments, Gemini-pro's F1 scores are higher than \approach
's by 1.74\%, 2.16\% and 4.36\%, respectively, indicating that Gemini-pro again has a slight edge over \approach in this aspect.

Then, we analyze the results of GPT4-V on a case in detail in Appendix~\ref{Appendix B: visual format recognition}.

\subsection{Performance of Spreadsheet Table Detection}
The results of GPT-4V and Gemini-pro for the spreadsheet table detection task are shown in Table 4. Firstly, we can see that the F1 scores obtained by having the VLMs decode the four boundaries and then applying our proposed mapping algorithm are significantly higher than those obtained by directly outputting the address range (e.g., "A1:C10"). Specifically, GPT-4V's zero-shot performance is 5.7\% and 14.38\% higher, and Gemini-pro's is 9.06\% and 11,86\% higher,respectively, which can be attributed to their excellent OCR capabilities. 

Secondly, both GPT-4V and Gemini-pro fall significantly short when compared to TableSense, with the closest F1 result still being 52.38\% lower. However, it is worth noting that TableSense inputs inputs serialized text from the spreadsheet, whereas the VLMs we are exploring take images as input. This indicates that there is a long way to go in continuously improving VLMs to achieve results comparable to text input.

Moreover, we observed an anomalous result: the one-shot results of \approach and Gemini-pro are generally worse than their zero-shot results. This might be due to the complex structure of spreadsheets, where providing an example can lead VLMs to favor outputs with structures similar to the example, resulting in misjudgments.

Finally, we analyze the results of GPT4-V on a case in detail in Appendix~\ref{Appendix B: spreadsheet table detection}.

\section{Conclusion and Future Work}

In this paper, we develop a suite of probing tasks aimed at evaluating the critical capabilities of VLMs in OCR, comprehension of formatting details, and recognition of spatial positioning within spreadsheet images. Our findings demonstrate that while VLMs possess strong OCR capabilities, they are prone to cell omission and prediction misalignment during OCR tasks on spreadsheet images. Furthermore, their spatial perception is insufficient, as they struggle to accurately determine the row and column numbers of cells in a two-dimensional spreadsheet grid. Surprisingly, VLMs cannot comprehend visual formats well like humans. Additionally, we introduce a spreadsheet table detection task designed to thoroughly assess the ability of VLMs to interpret spreadsheet images effectively. However, the performance of this task falls short of that achieved by existing SOTA method, indicating that processing and comprehending spreadsheets remains a significant challenge. 

Future research could focus on handling larger spreadsheet images and segmenting these spreadsheets without compromising the integrity of their format and spatial relationships. Despite these challenges, the potential benefits of treating spreadsheets as images are substantial. In this paper, we have proposed methods that can massively generate spreadsheet-image pairs, and under our proposed settings, we can control various challenges. Utilizing these methods to generate large amounts of data, we train open-source large models to enhance their understanding of structured data on grids, further advancing the comprehensive capabilities of vision language models.



\bibliography{main}

\begin{thebibliography}{28}
\providecommand{\natexlab}[1]{#1}

\bibitem[{Achiam et~al.(2023)Achiam, Adler, Agarwal, Ahmad, Akkaya, Aleman, Almeida, Altenschmidt, Altman, Anadkat et~al.}]{achiam2023gpt}
Josh Achiam, Steven Adler, Sandhini Agarwal, Lama Ahmad, Ilge Akkaya, Florencia~Leoni Aleman, Diogo Almeida, Janko Altenschmidt, Sam Altman, Shyamal Anadkat, et~al. 2023.
\newblock Gpt-4 technical report.
\newblock \emph{arXiv preprint arXiv:2303.08774}.

\bibitem[{Birch et~al.(2018)Birch, Lyford-Smith, and Guo}]{birch2018future}
David Birch, David Lyford-Smith, and Yike Guo. 2018.
\newblock The future of spreadsheets in the big data era.
\newblock \emph{arXiv preprint arXiv:1801.10231}.

\bibitem[{Chen et~al.(2020)Chen, Wang, Chen, Zhang, Wang, Li, Zhou, and Wang}]{chen2020tabfact}
Wenhu Chen, Hongmin Wang, Jianshu Chen, Yunkai Zhang, Hong Wang, Shiyang Li, Xiyou Zhou, and William~Yang Wang. 2020.
\newblock \href {https://arxiv.org/abs/1909.02164} {Tabfact: A large-scale dataset for table-based fact verification}.
\newblock \emph{Preprint}, arXiv:1909.02164.

\bibitem[{Chen et~al.(2024)Chen, Yuan, Zhang, Zheng, Liu, Ni, and Hao}]{chen2024sheetagent}
Yibin Chen, Yifu Yuan, Zeyu Zhang, Yan Zheng, Jinyi Liu, Fei Ni, and Jianye Hao. 2024.
\newblock Sheetagent: A generalist agent for spreadsheet reasoning and manipulation via large language models.
\newblock \emph{arXiv preprint arXiv:2403.03636}.

\bibitem[{Deng et~al.(2024)Deng, Sun, He, Sikka, Chen, Ma, Zhang, and Mihalcea}]{deng2024tables}
Naihao Deng, Zhenjie Sun, Ruiqi He, Aman Sikka, Yulong Chen, Lin Ma, Yue Zhang, and Rada Mihalcea. 2024.
\newblock Tables as images? exploring the strengths and limitations of llms on multimodal representations of tabular data.
\newblock \emph{arXiv preprint arXiv:2402.12424}.

\bibitem[{Dong et~al.(2022)Dong, Cheng, He, Zhou, Zhou, Zhou, Liu, Han, and Zhang}]{dong2022table}
Haoyu Dong, Zhoujun Cheng, Xinyi He, Mengyu Zhou, Anda Zhou, Fan Zhou, Ao~Liu, Shi Han, and Dongmei Zhang. 2022.
\newblock Table {P}retraining: A survey on model architectures, pretraining objectives, and downstream tasks.
\newblock \emph{Proceedings of the Thirty-First International Joint Conference on Artificial Intelligence}.

\bibitem[{Dong et~al.(2019)Dong, Liu, Han, Fu, and Zhang}]{dong2019tablesense}
Haoyu Dong, Shijie Liu, Shi Han, Zhouyu Fu, and Dongmei Zhang. 2019.
\newblock Tablesense: Spreadsheet table detection with convolutional neural networks.
\newblock In \emph{Proceedings of the AAAI conference on artificial intelligence}, volume~33, pages 69--76.

\bibitem[{Dong and Wang(2024)}]{dong2024large}
Haoyu Dong and Zhiruo Wang. 2024.
\newblock Large language models for tabular data: Progresses and future directions.
\newblock In \emph{Proceedings of the 47th International ACM SIGIR Conference on Research and Development in Information Retrieval}, pages 2997--3000.

\bibitem[{Fang et~al.(2012)Fang, Mitra, Tang, and Giles}]{fang2012table}
Jing Fang, Prasenjit Mitra, Zhi Tang, and C~Lee Giles. 2012.
\newblock Table header detection and classification.
\newblock In \emph{Proceedings of the AAAI Conference on Artificial Intelligence}, volume~26, pages 599--605.

\bibitem[{Guo et~al.(2023)Guo, Du, and Liu}]{guo2023gpt4graph}
Jiayan Guo, Lun Du, and Hengyu Liu. 2023.
\newblock Gpt4graph: Can large language models understand graph structured data? an empirical evaluation and benchmarking.
\newblock \emph{arXiv preprint arXiv:2305.15066}.

\bibitem[{Huang et~al.(2023)Huang, Lu, Chen, Li, Xie, Zhu, Gao, and Peng}]{huang2023improving}
Yongshuai Huang, Ning Lu, Dapeng Chen, Yibo Li, Zecheng Xie, Shenggao Zhu, Liangcai Gao, and Wei Peng. 2023.
\newblock Improving table structure recognition with visual-alignment sequential coordinate modeling.
\newblock In \emph{Proceedings of the IEEE/CVF Conference on Computer Vision and Pattern Recognition}, pages 11134--11143.

\bibitem[{Jiang et~al.(2023)Jiang, Zhou, Dong, Ye, Zhao, and Wen}]{jiang2023structgpt}
Jinhao Jiang, Kun Zhou, Zican Dong, Keming Ye, Wayne~Xin Zhao, and Ji-Rong Wen. 2023.
\newblock Structgpt: A general framework for large language model to reason over structured data.
\newblock \emph{arXiv preprint arXiv:2305.09645}.

\bibitem[{Lauren{\c{c}}on et~al.(2024)Lauren{\c{c}}on, Tronchon, Cord, and Sanh}]{laurenccon2024matters}
Hugo Lauren{\c{c}}on, L{\'e}o Tronchon, Matthieu Cord, and Victor Sanh. 2024.
\newblock What matters when building vision-language models?
\newblock \emph{arXiv preprint arXiv:2405.02246}.

\bibitem[{Li et~al.(2024)Li, Su, Chen, Li, and ZHANG}]{li2024sheetcopilot}
Hongxin Li, Jingran Su, Yuntao Chen, Qing Li, and ZHAO-XIANG ZHANG. 2024.
\newblock Sheetcopilot: Bringing software productivity to the next level through large language models.
\newblock \emph{Advances in Neural Information Processing Systems}, 36.

\bibitem[{Li et~al.(2023)Li, He, Yashar, Cui, Ge, Zhang, Fainman, Zhang, and Chaudhuri}]{li2023table}
Peng Li, Yeye He, Dror Yashar, Weiwei Cui, Song Ge, Haidong Zhang, Danielle~Rifinski Fainman, Dongmei Zhang, and Surajit Chaudhuri. 2023.
\newblock Table-gpt: Table-tuned gpt for diverse table tasks.
\newblock \emph{arXiv preprint arXiv:2310.09263}.

\bibitem[{Ly and Takasu(2023)}]{ly2023end}
Nam~Tuan Ly and Atsuhiro Takasu. 2023.
\newblock An end-to-end multi-task learning model for image-based table recognition.
\newblock \emph{arXiv preprint arXiv:2303.08648}.

\bibitem[{Nassar et~al.()Nassar, Livathinos, Lysak, and Staar}]{nassartableformer}
A~Nassar, N~Livathinos, M~Lysak, and PWJ Staar.
\newblock Tableformer: table structure understanding with transformers. corr abs/2203.01017 (2022).

\bibitem[{Novikova et~al.(2017)Novikova, Dušek, and Rieser}]{novikova2017e2e}
Jekaterina Novikova, Ondřej Dušek, and Verena Rieser. 2017.
\newblock \href {https://arxiv.org/abs/1706.09254} {The e2e dataset: New challenges for end-to-end generation}.
\newblock \emph{Preprint}, arXiv:1706.09254.

\bibitem[{Pasupat and Liang(2015)}]{pasupat2015compositional}
Panupong Pasupat and Percy Liang. 2015.
\newblock \href {https://arxiv.org/abs/1508.00305} {Compositional semantic parsing on semi-structured tables}.
\newblock \emph{Preprint}, arXiv:1508.00305.

\bibitem[{Singh et~al.(2023)Singh, Cambronero, Gulwani, Le, and Verbruggen}]{singh2023assessing}
Mukul Singh, Jos{\'e} Cambronero, Sumit Gulwani, Vu~Le, and Gust Verbruggen. 2023.
\newblock Assessing gpt4-v on structured reasoning tasks.
\newblock \emph{arXiv preprint arXiv:2312.11524}.

\bibitem[{Singha et~al.(2023)Singha, Cambronero, Gulwani, Le, and Parnin}]{singha2023tabular}
Ananya Singha, Jos{\'e} Cambronero, Sumit Gulwani, Vu~Le, and Chris Parnin. 2023.
\newblock Tabular representation, noisy operators, and impacts on table structure understanding tasks in llms.
\newblock \emph{arXiv preprint arXiv:2310.10358}.

\bibitem[{Sui et~al.(2023{\natexlab{a}})Sui, Zhou, Zhou, Han, and Zhang}]{sui2023gpt4table}
Yuan Sui, Mengyu Zhou, Mingjie Zhou, Shi Han, and Dongmei Zhang. 2023{\natexlab{a}}.
\newblock Gpt4table: Can large language models understand structured table data? a benchmark and empirical study.
\newblock \emph{arXiv preprint ArXiv:2305.13062}.

\bibitem[{Sui et~al.(2023{\natexlab{b}})Sui, Zou, Zhou, He, Du, Han, and Zhang}]{sui2023tap4llm}
Yuan Sui, Jiaru Zou, Mengyu Zhou, Xinyi He, Lun Du, Shi Han, and Dongmei Zhang. 2023{\natexlab{b}}.
\newblock Tap4llm: Table provider on sampling, augmenting, and packing semi-structured data for large language model reasoning.
\newblock \emph{arXiv preprint arXiv:2312.09039}.

\bibitem[{Tang et~al.(2023)Tang, Zong, Zhao, Cohan, and Gerstein}]{tang2023struc}
Xiangru Tang, Yiming Zong, Yilun Zhao, Arman Cohan, and Mark Gerstein. 2023.
\newblock Struc-bench: Are large language models really good at generating complex structured data?
\newblock \emph{arXiv preprint arXiv:2309.08963}.

\bibitem[{Team et~al.(2023)Team, Anil, Borgeaud, Wu, Alayrac, Yu, Soricut, Schalkwyk, Dai, Hauth et~al.}]{team2023gemini}
Gemini Team, Rohan Anil, Sebastian Borgeaud, Yonghui Wu, Jean-Baptiste Alayrac, Jiahui Yu, Radu Soricut, Johan Schalkwyk, Andrew~M Dai, Anja Hauth, et~al. 2023.
\newblock Gemini: a family of highly capable multimodal models.
\newblock \emph{arXiv preprint arXiv:2312.11805}.

\bibitem[{Wang et~al.(2021)Wang, Dong, Jia, Li, Fu, Han, and Zhang}]{wang2021tuta}
Zhiruo Wang, Haoyu Dong, Ran Jia, Jia Li, Zhiyi Fu, Shi Han, and Dongmei Zhang. 2021.
\newblock Tuta: Tree-based transformers for generally structured table pre-training.
\newblock In \emph{Proceedings of the 27th ACM SIGKDD Conference on Knowledge Discovery \& Data Mining}, pages 1780--1790.

\bibitem[{Wu et~al.(2023)Wu, Chen, Bu, Ji, Zhang, and Sheng}]{wu2023huss}
Xindong Wu, Hao Chen, Chenyang Bu, Shengwei Ji, Zan Zhang, and Victor~S Sheng. 2023.
\newblock Huss: A heuristic method for understanding the semantic structure of spreadsheets.
\newblock \emph{Data Intelligence}, 5(3):537--559.

\bibitem[{Zhang et~al.(2023)Zhang, Yue, Li, and Sun}]{zhang2023tablellama}
Tianshu Zhang, Xiang Yue, Yifei Li, and Huan Sun. 2023.
\newblock Tablellama: Towards open large generalist models for tables.
\newblock \emph{arXiv preprint arXiv:2311.09206}.

\end{thebibliography}

\clearpage
\newpage

\appendix
\section{Prompt Examples}
\label{appendix: A}

\begin{figure}[h]
    \centering
    \includegraphics[width=\linewidth]{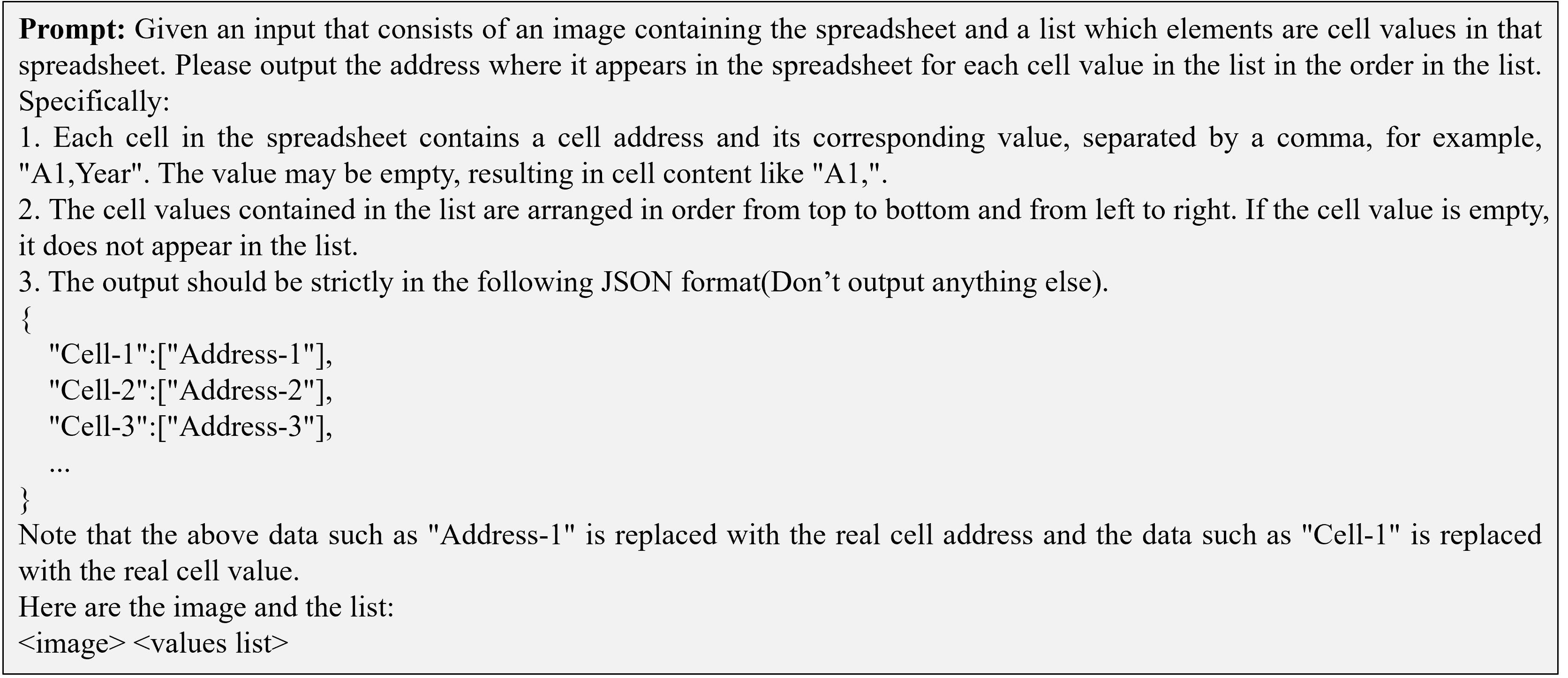}
    \caption{The prompt for address augment experiments of spatial position perception task.}
    \label{fig:address_prompt}
\end{figure}

\begin{figure}[h]
    \centering
    \includegraphics[width=\linewidth]{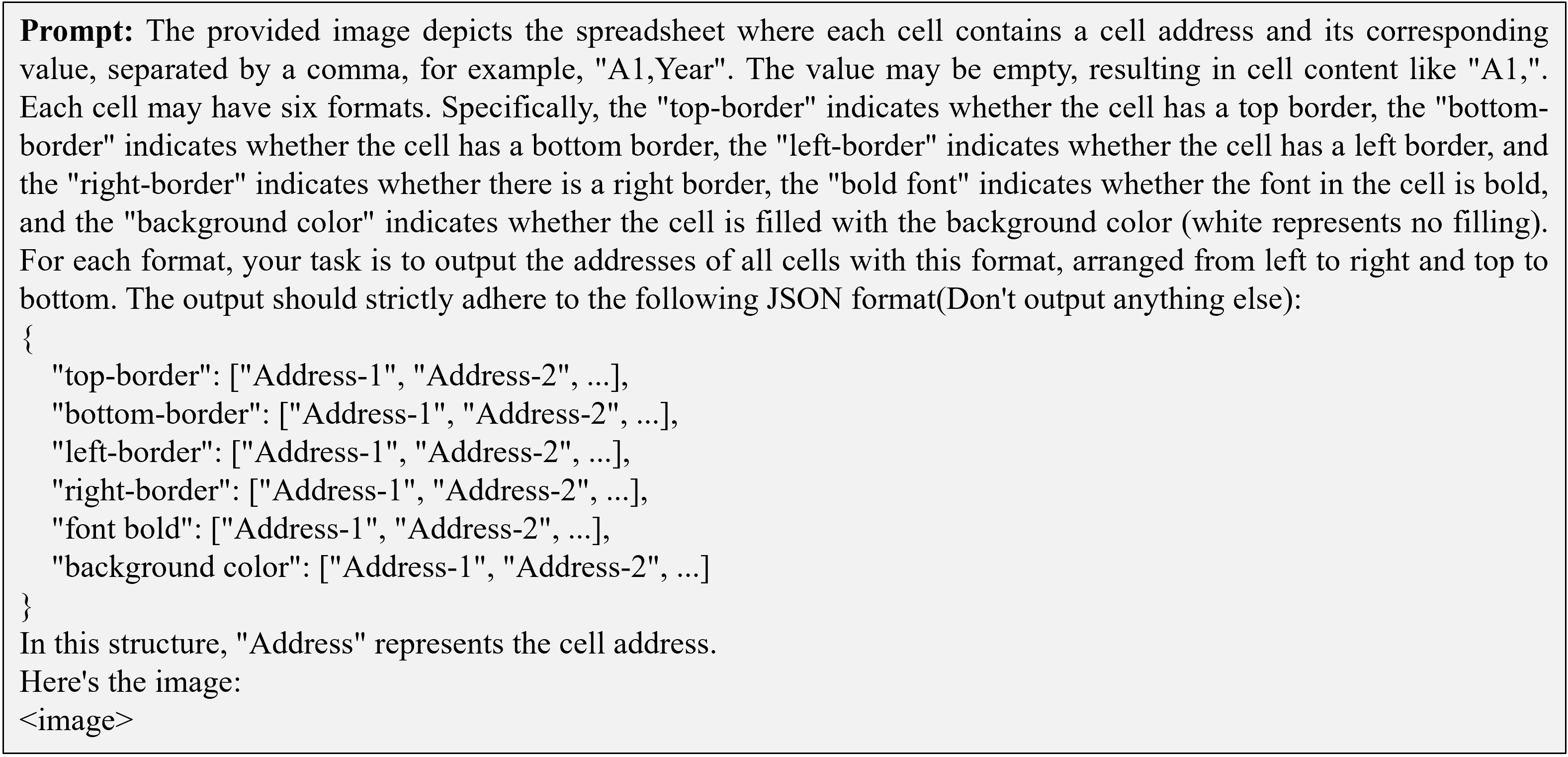}
    \caption{The prompt for address augment experiments of visual format recognition task.}
    \label{fig:format_prompt}
\end{figure}

\begin{figure}[ht]
    \centering
    \begin{subfigure}{0.48\textwidth}
        \includegraphics[width=\linewidth]{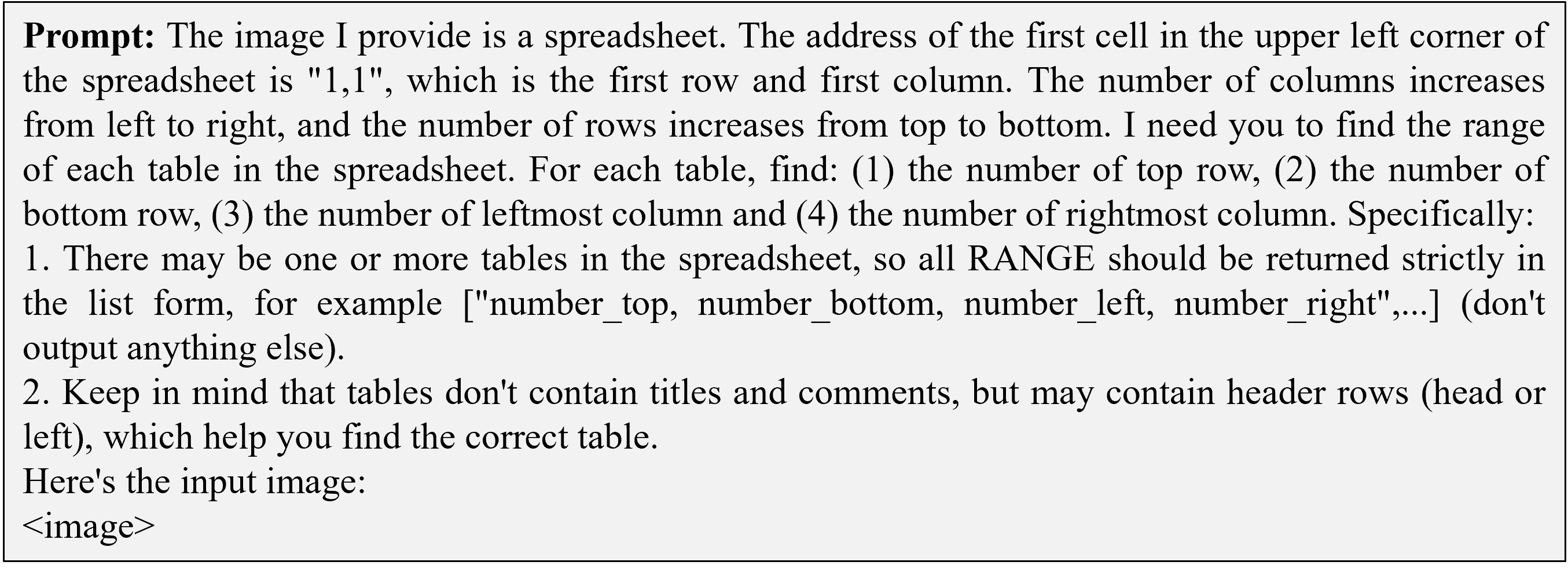}
        \caption{The zero-shot prompt for outputting ranges in vanilla experiments.}
        \label{fig:TD_prompt_all_1}
    \end{subfigure}
        \begin{subfigure}{0.48\textwidth}
        \includegraphics[width=\linewidth]{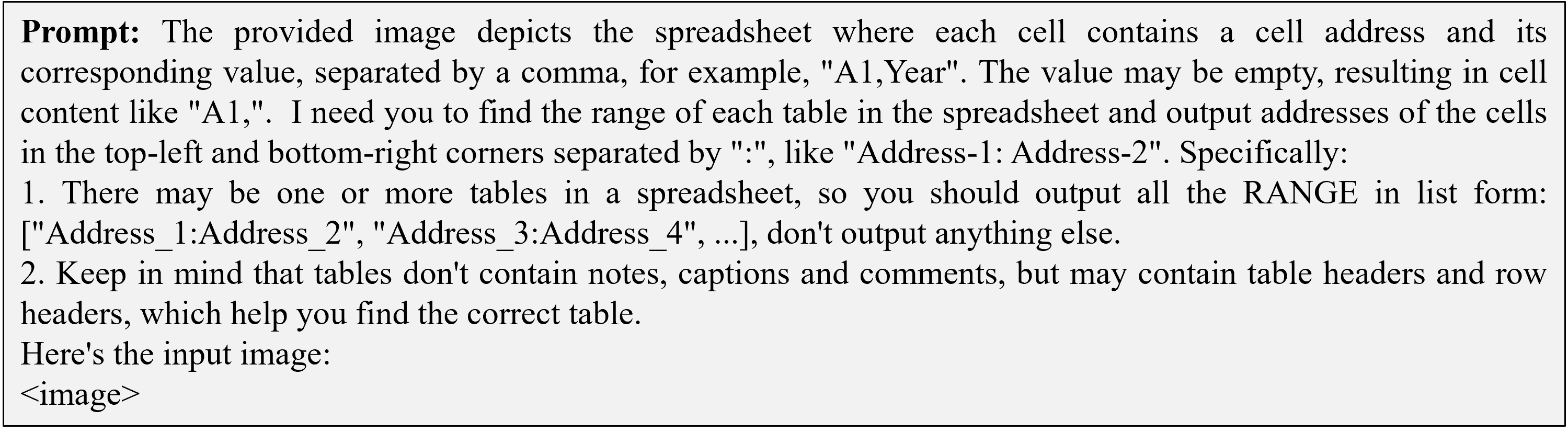}
        \caption{The zero-shot prompt for outputting ranges in address augment experiments.}
        \label{fig:TD_prompt_all_2}
    \end{subfigure}

    \begin{subfigure}{0.48\textwidth}
        \includegraphics[width=\linewidth]{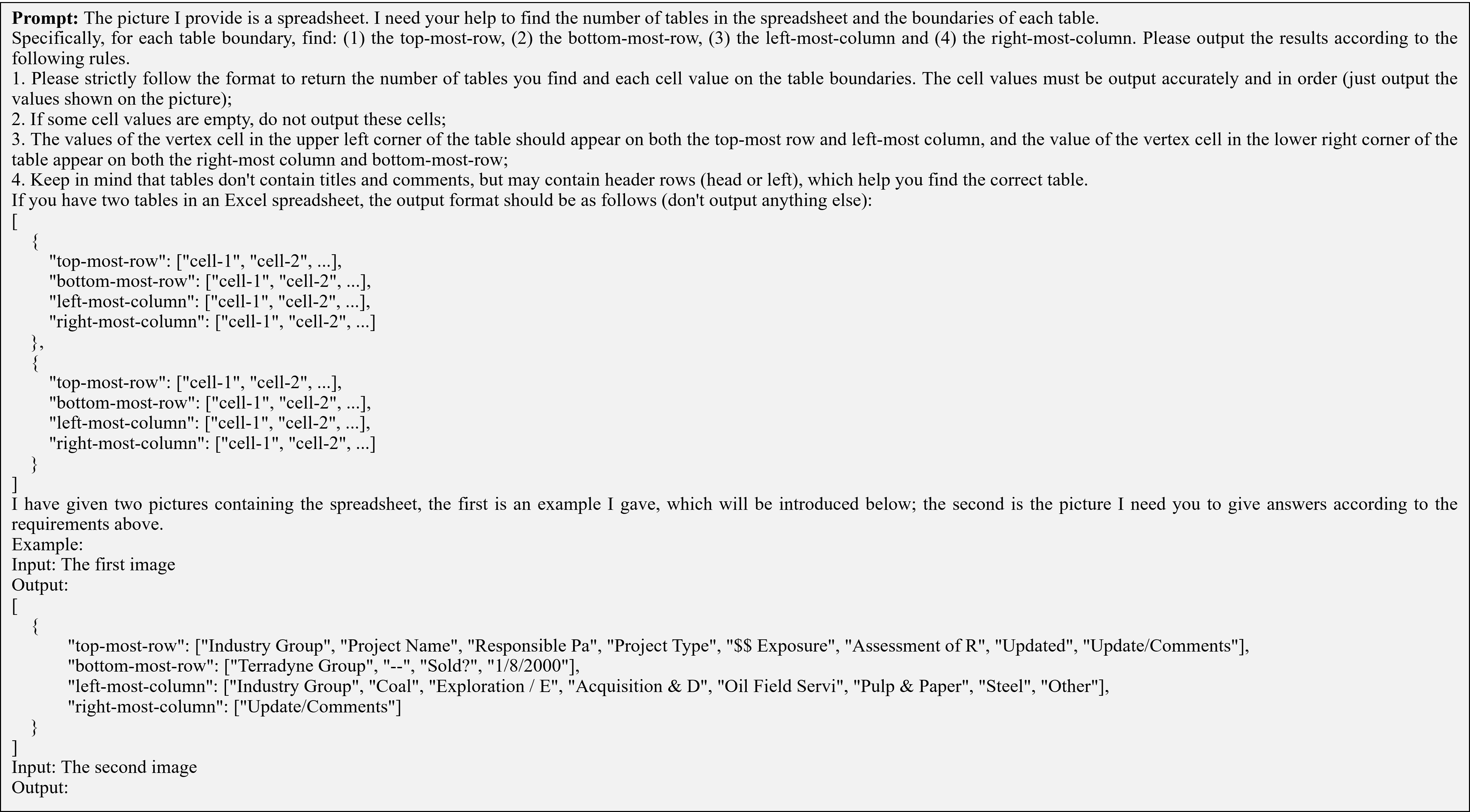}
        \caption{The one-shot prompt for decoding four boundaries in vanilla experiments}
        \label{fig:TD_prompt_all_3}
    \end{subfigure}
 
    \begin{subfigure}{0.48\textwidth}
        \includegraphics[width=\linewidth]{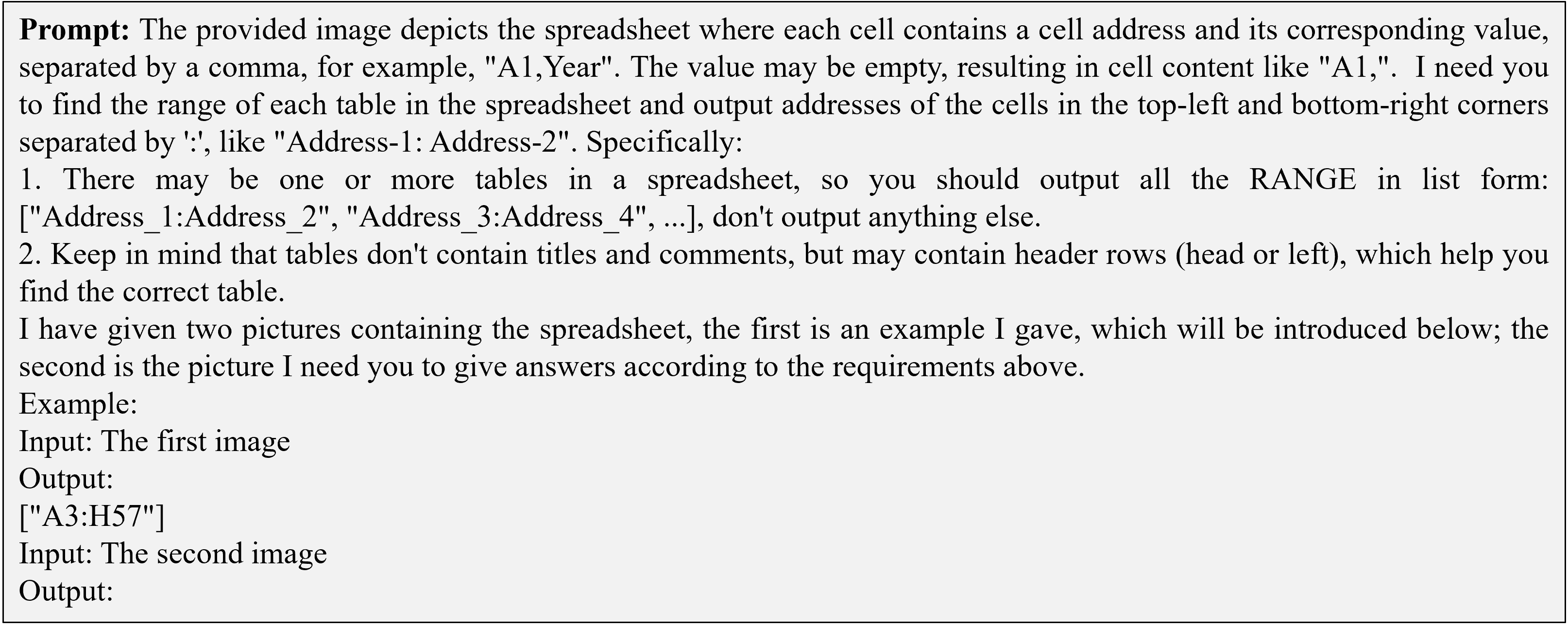}
        \caption{The one-shot prompt for outputting ranges in address augment experiments.}
        \label{fig:TD_prompt_all_4}
    \end{subfigure}
    
    \caption{The prompt of spreadsheet table detection task.}
    \label{fig:TD_prompt_all}
\end{figure}

Figure.~\ref{fig:address_prompt} shows the prompt for address augment experiment in spatial position perception task;  Figure.~\ref{fig:format_prompt} shows the prompt for address augment experiment in visual format recognition task; Figure.~\ref{fig:TD_prompt_all} shows the prompt for spreadsheet table detection task under different experiment setting and output format.

\section{Case Study}

\begin{figure}[h]
    \centering
    \begin{subfigure}{0.5\textwidth}
        \includegraphics[width=\linewidth]{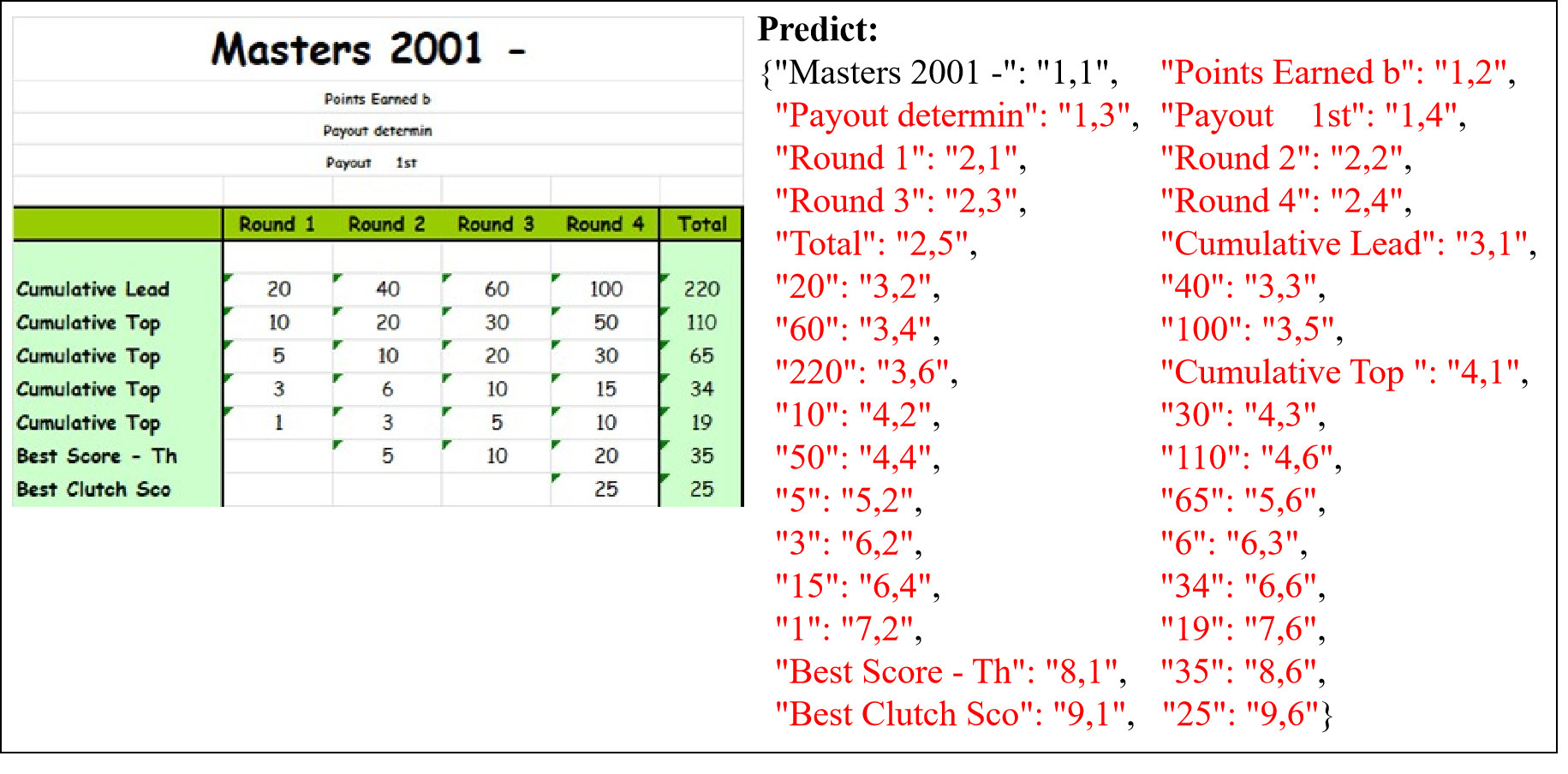}
        \caption{Vanilla}
        \label{fig:case_study_2-a}
    \end{subfigure}
    \begin{subfigure}{0.5\textwidth}
        \includegraphics[width=\linewidth]{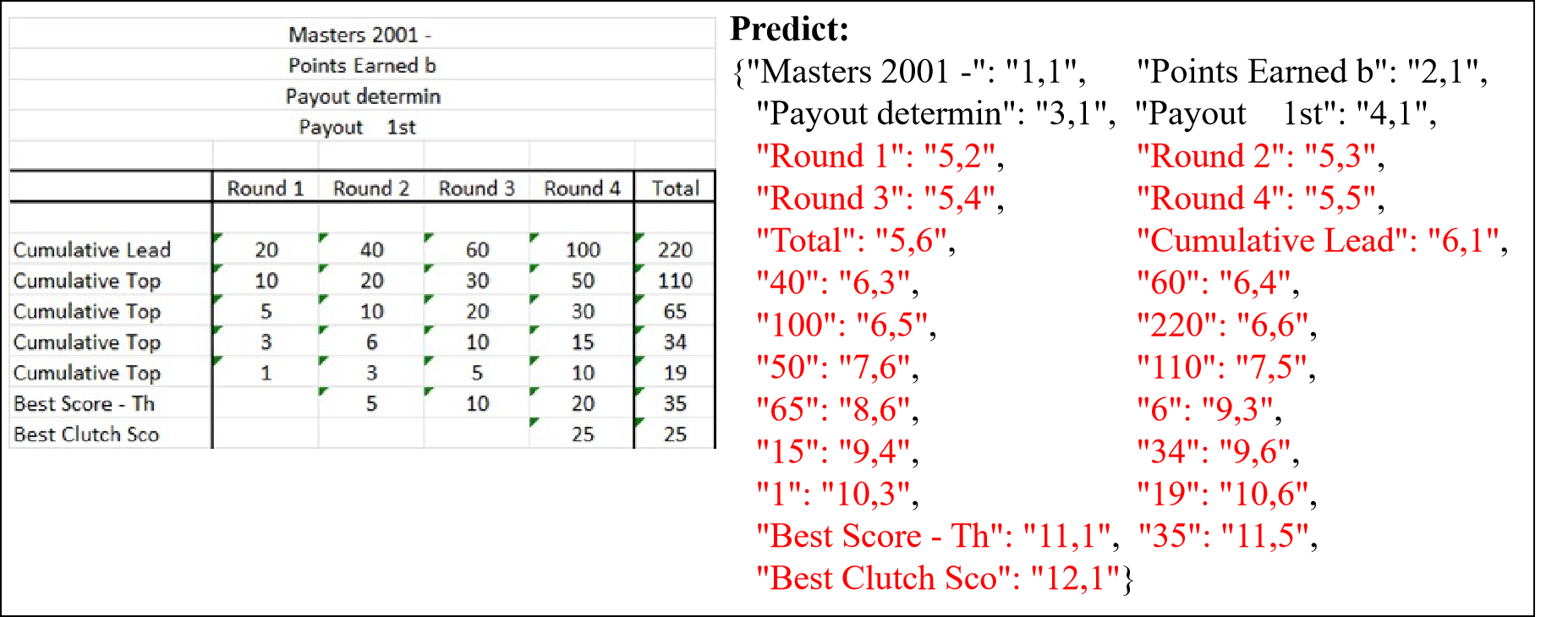}
        \caption{Style Change}
        \label{fig:case_study_2-b}
    \end{subfigure}
    \begin{subfigure}{0.5\textwidth}
        \includegraphics[width=\linewidth]{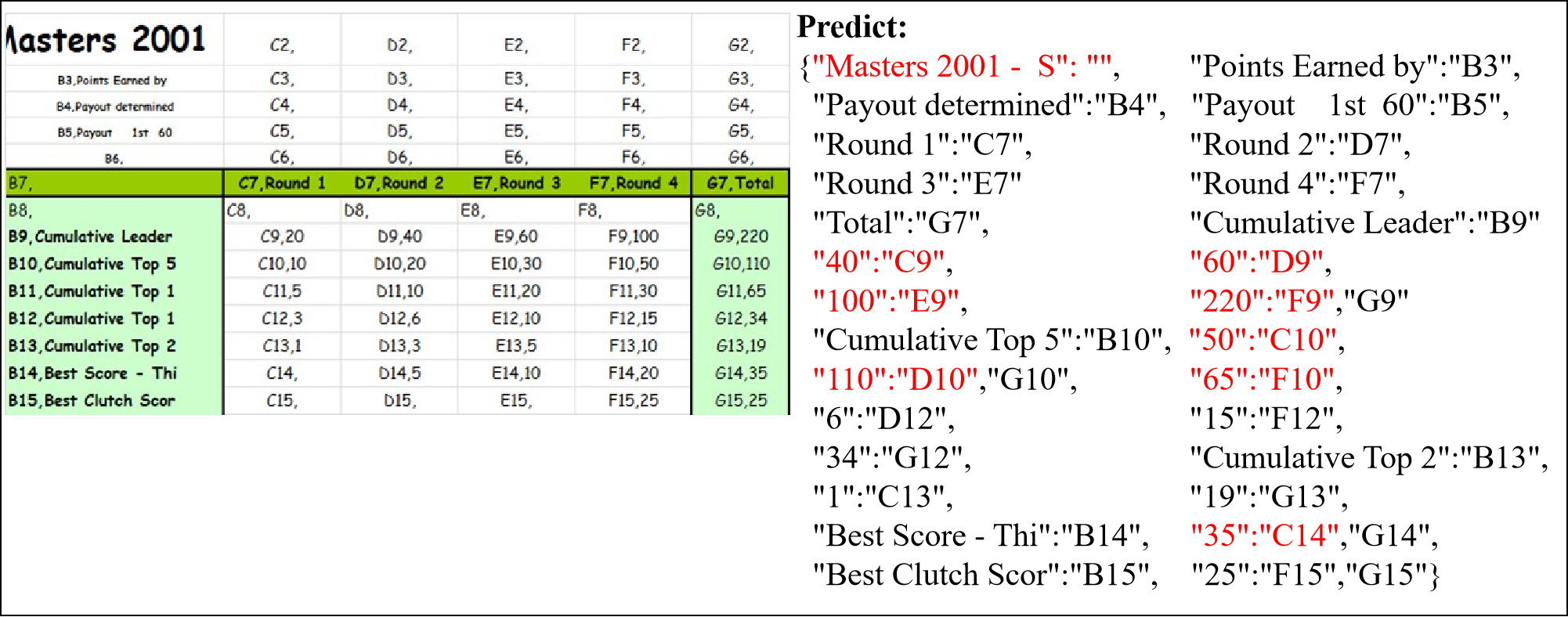}
        \caption{Address Augment}
        \label{fig:case_study_2-c}
    \end{subfigure}
    \caption{An example of GPT-4V on spatial position perception task. The content marked in red indicates
the LCS match error prediction}
    \label{fig:case_study_2}
\end{figure}

\begin{figure}[h]
    \centering
    \begin{subfigure}{0.5\textwidth}
        \includegraphics[width=\linewidth]{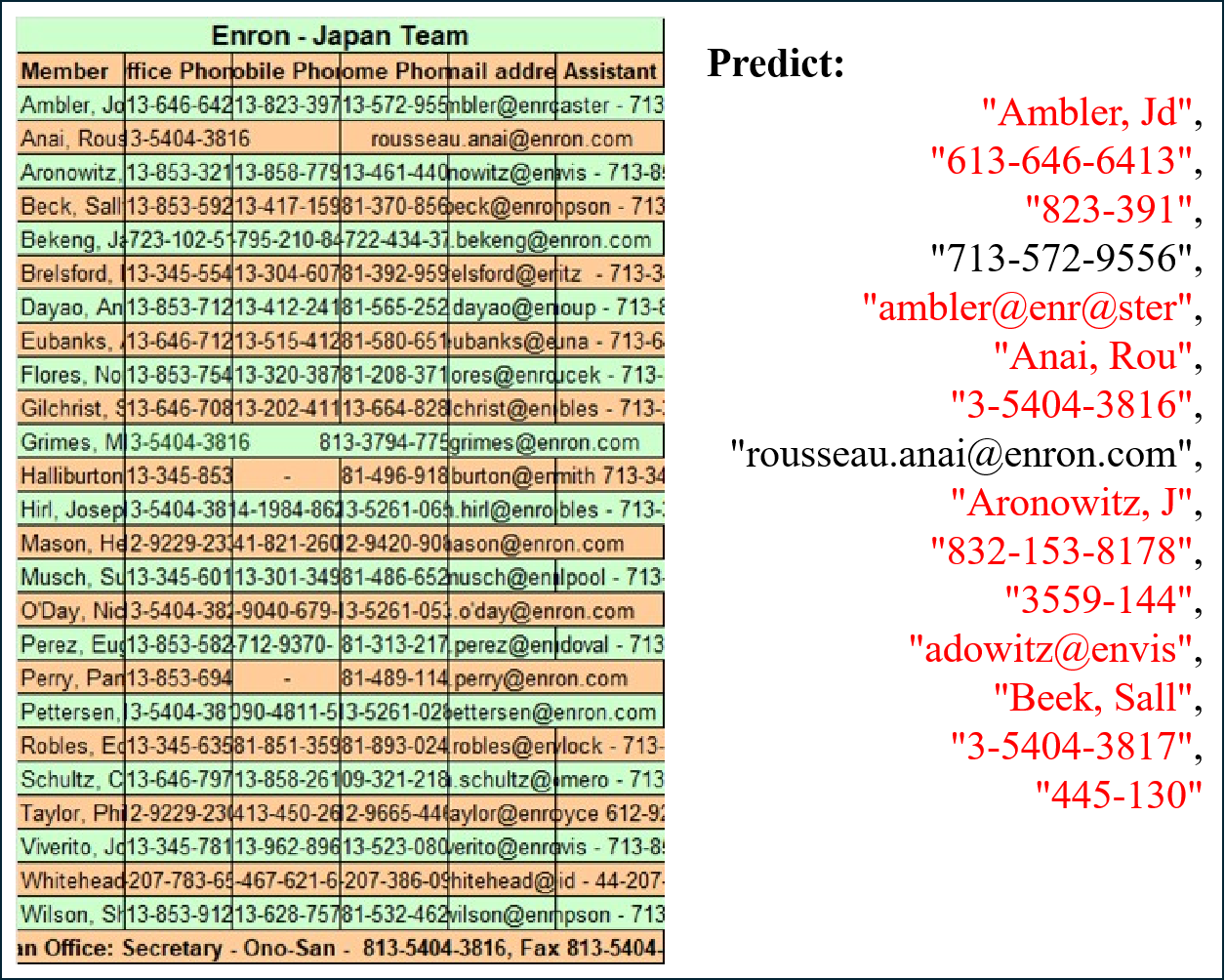}
        \caption{Vanilla}
        \label{fig:case_study_1-a}
    \end{subfigure}
    \begin{subfigure}{0.5\textwidth}
        \includegraphics[width=\linewidth]{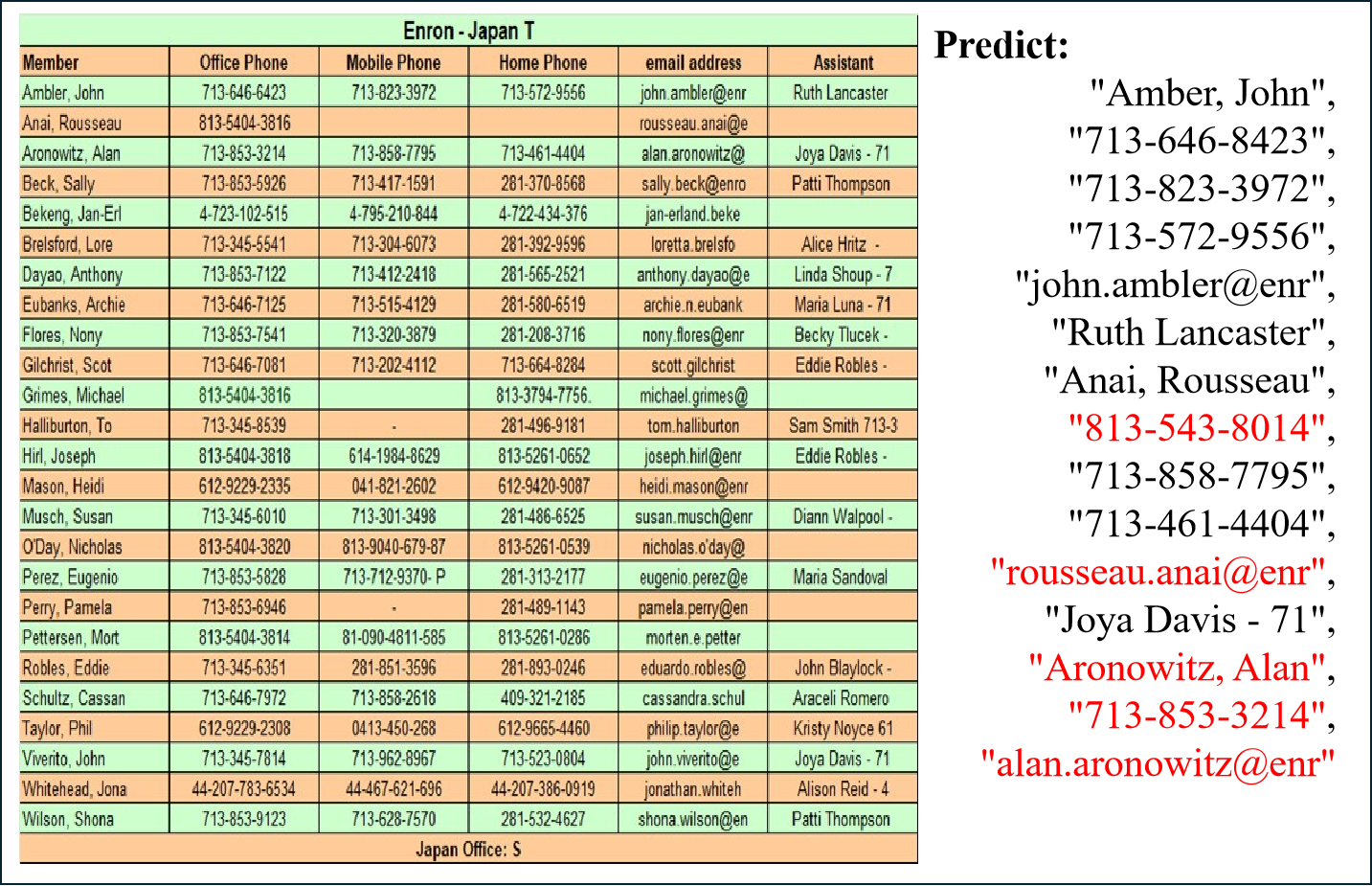}
        \caption{ColWidth Adjust}
        \label{fig:case_study_1-b}
    \end{subfigure}
    \begin{subfigure}{0.5\textwidth}
        \includegraphics[width=\linewidth]{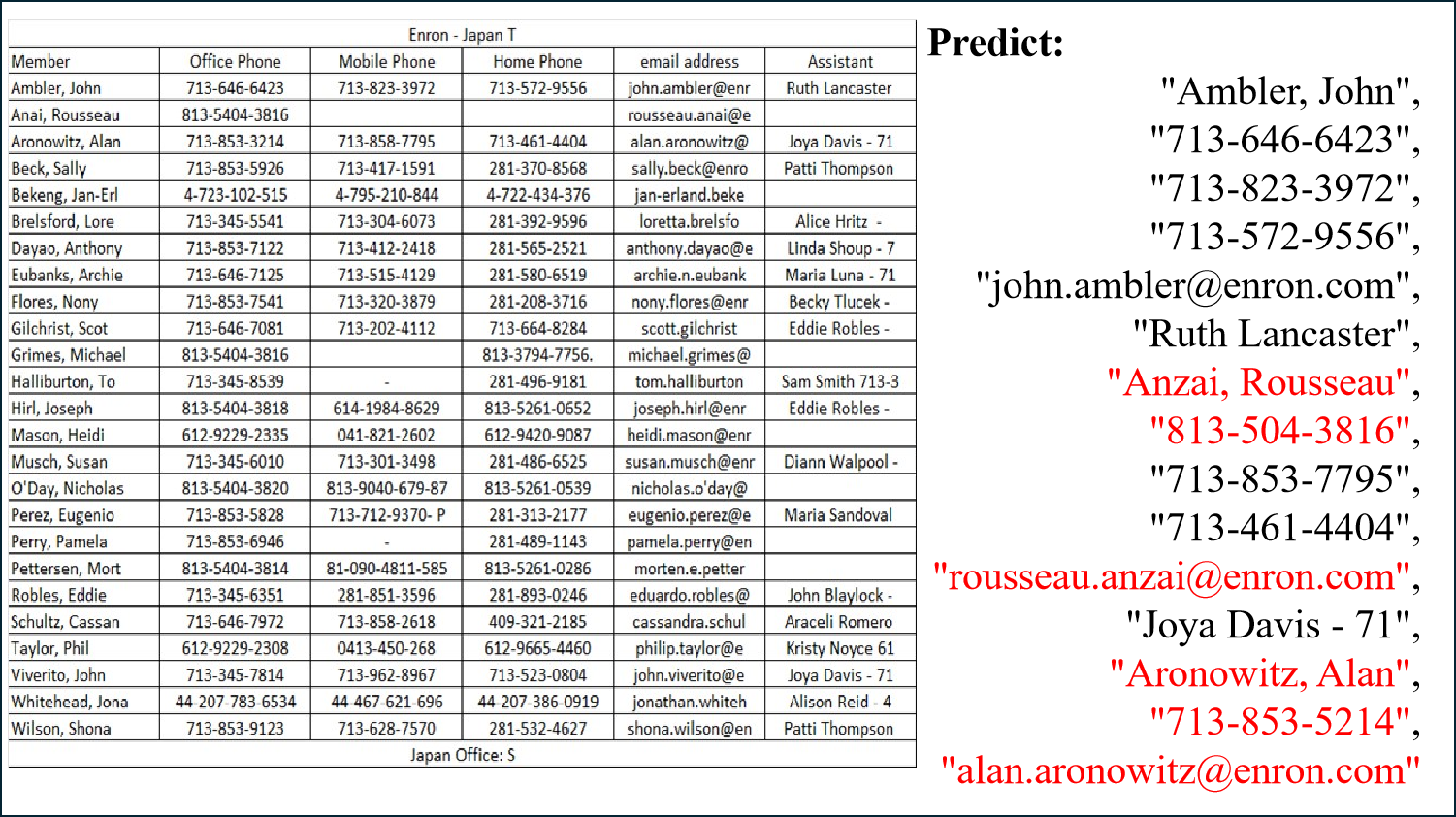}
        \caption{Style Change}
        \label{fig:case_study_1-c}
    \end{subfigure}
    \caption{An example of GPT-4V on OCR task. Due to space limitations, only the contents of the first 15 cells are shown. The content marked in red indicates the LCS match error prediction.}
    \label{fig:case_study_1}
\end{figure}

\subsection{A case of OCR task}
\label{Appendix B: ocr}
In order to deeply explore the impact of different reconstructed spreadsheet images on \approach's OCR capabilities, we will analyze the case shown in Figure.~\ref{fig:case_study_1} in detail.

First, by comparing the results of Figure.~\ref{fig:case_study_1-a} and Figure.~\ref{fig:case_study_1-b}, we can clearly find that not adjusting the column width in the spreadsheet will cause the OCR capability of \approach to drop significantly. This is due to the fact that the cell content in many spreadsheets will not be fully displayed when the column width is not adjusted, and there may be overlap or coverage between adjacent cells, as shown in Figure.~\ref{fig:case_study_1-a}.

Secondly, by observing these three pictures, we will find that \approach has insufficient positioning capabilities when performing OCR, resulting in some cells being missed or misplaced during the prediction process. For example, \approach's prediction results for the three pictures in Figure.~\ref{fig:case_study_1} ignore the first two lines of the spreadsheet, and in both Figure.~\ref{fig:case_study_1-b} and Figure.~\ref{fig:case_study_1-c}, \approach predicts "Aronowitz, Alan" to "713-858-7795" After, but actually it should be in front.

\subsection{A case of spatial position perception task}
\label{Appendix B: spatial position perception}

Figure.~\ref{fig:case_study_2} presents a tangible example evaluating \approach's proficiency in spatial position awareness within spreadsheet environments. Upon scrutiny, it's apparent that even in relatively straightforward scenarios, both vanilla and style change experiments reveal \approach's inadequate performance in accurately predicting position. While \approach effectively forecasts column positions for most cells, it consistently struggles with row positions, consistently displaying an offset. This issue becomes more pronounced in the presence of empty rows, leading to inaccuracies in subsequent cell position predictions.

In contrast, the address augment experiment showcases a comparatively better performance by \approach. This improvement can be attributed to its impressive OCR capabilities, allowing it to accurately identify and pair cell addresses with their corresponding values within a single cell.

\begin{figure}[h]
    \centering
    \begin{subfigure}{0.45\textwidth}
        \includegraphics[width=\linewidth]{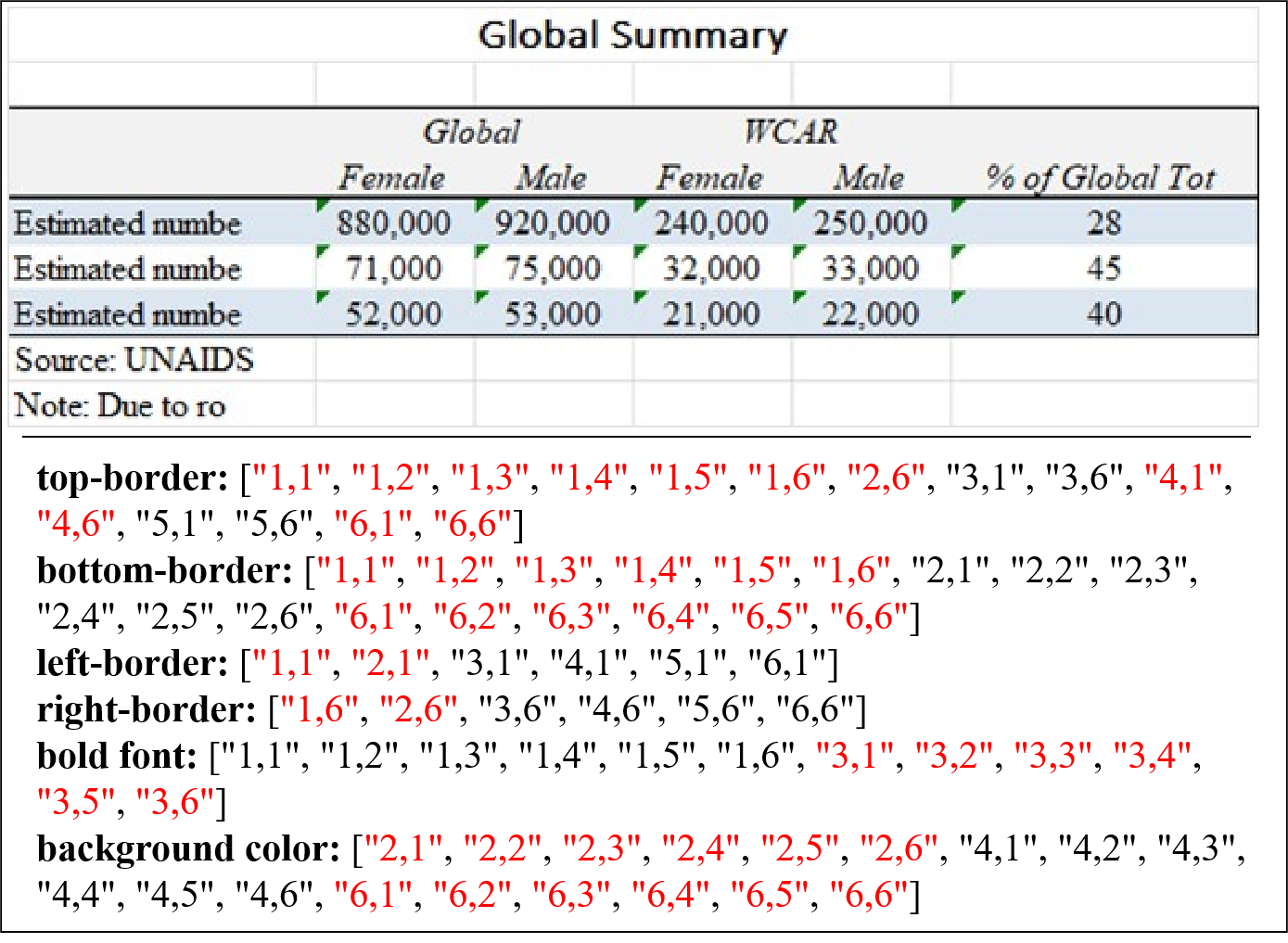}
        \caption{Vanilla}
        \label{fig:case_study_3-a}
    \end{subfigure}
    \begin{subfigure}{0.45\textwidth}
        \includegraphics[width=\linewidth]{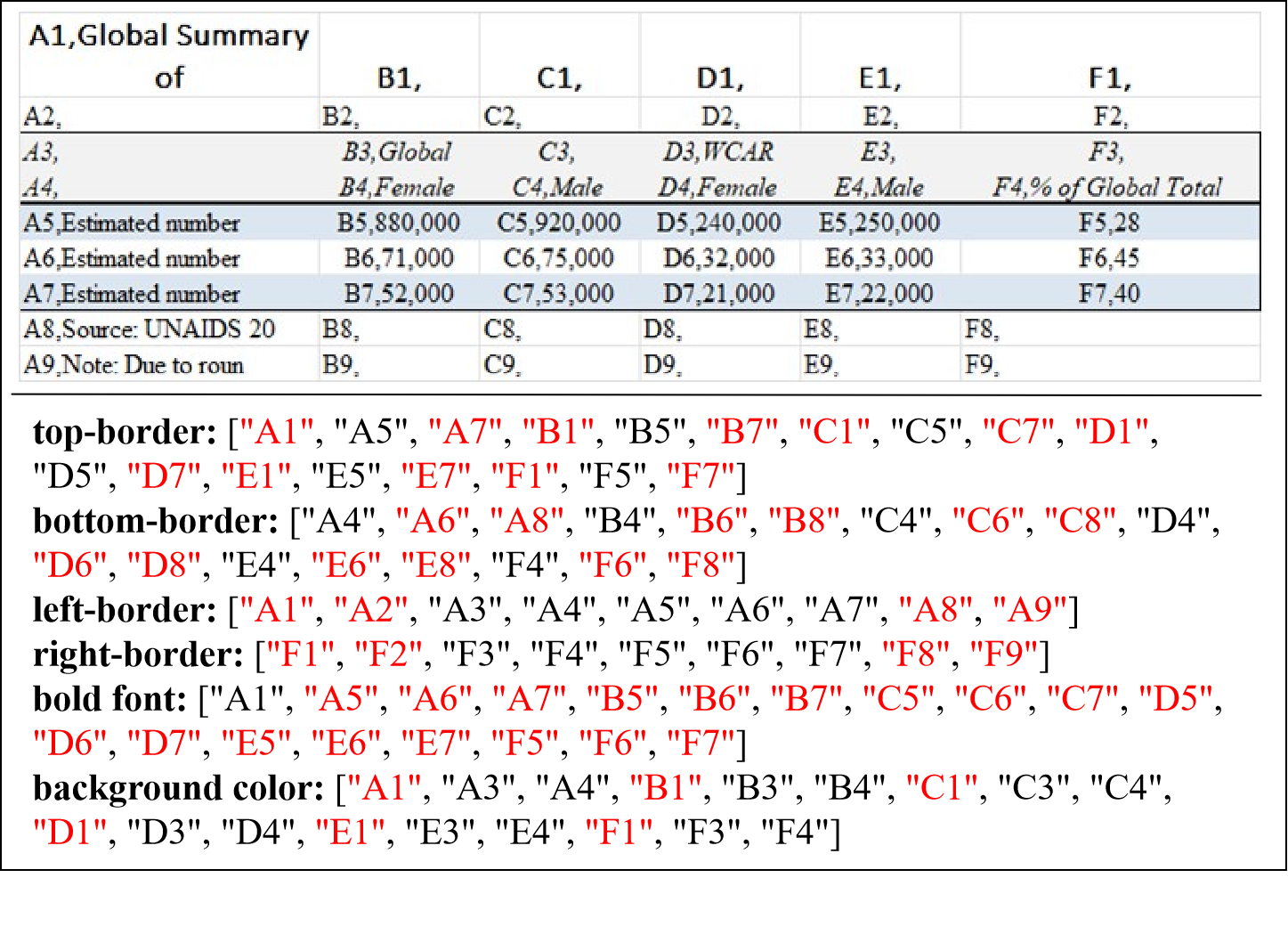}
        \caption{Address Augment}
        \label{fig:case_study_3-b}
    \end{subfigure}
    \caption{An example of GPT-4V on visual format recognition task. The content marked in red indicates the wrong predictions.}
    \label{fig:case_study_3}
\end{figure}

\subsection{A case of visual format recognition task}
\label{Appendix B: visual format recognition}

The case depicted in Figure.~\ref{fig:case_study_3} presents the outcomes of \approach's analysis under vanilla and address augment experiments. Examination of these results reveals \approach's limited grasp of format information in both scenarios, indicating its potential inability to comprehend spreadsheet format details akin to humans, likely due to image encoding constraints. Upon meticulous scrutiny of \approach's outputs, a discernible trend emerges: it tends to follow imaginary rules to identify locations featuring specific formats. For instance, under the vanilla experiment, \approach consistently identifies the three-line area spanning from "1,1: 1,6", "2,1:2,6", and "6,1:6,6" for bottom borders. Similarly, under the address agument condition, it consistently outputs areas such as "A1:F1", "A5:F5", and "A7:F7" representing top borders.

\subsection{A case of spreadsheet table detection task}
\label{Appendix B: spreadsheet table detection}

In order to explore the performance of \approach on the spreadsheet table detection task, We will analyze the case in detail.

\begin{figure}[h]
    \centering
    \begin{subfigure}{0.45\textwidth}
        \includegraphics[width=\linewidth]{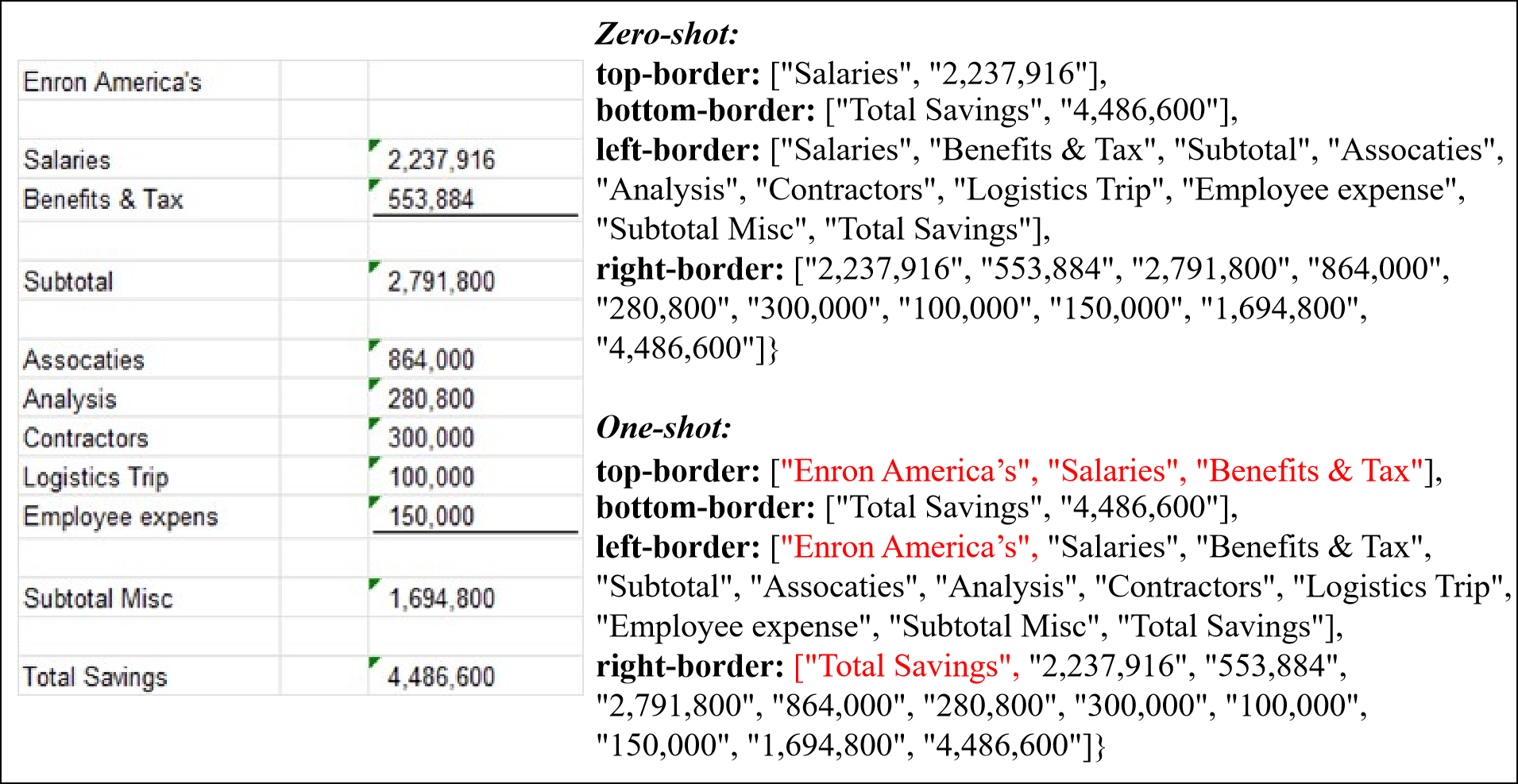}
        \caption{zeroshot vs. oneshot}
        \label{fig:case_study_4-a}
    \end{subfigure}
    \begin{subfigure}{0.45\textwidth}
        \includegraphics[width=\linewidth]{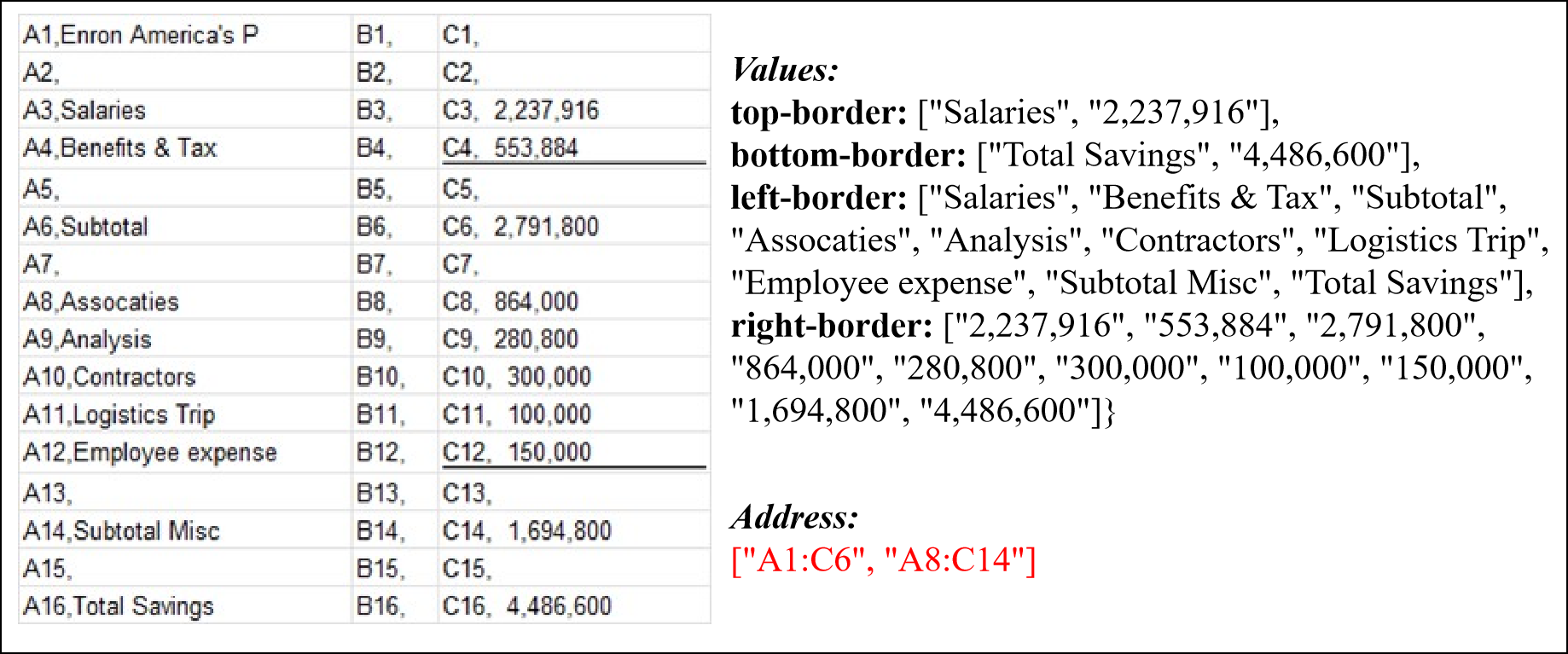}
        \caption{four vs. range}
        \label{fig:case_study_4-b}
    \end{subfigure}
    \caption{An example of GPT-4V on spreadsheet table detection task. The red color represents the wrong predictions.}
    \label{fig:case_study_3}
\end{figure}

First, by analyzing Figure.~\ref{fig:case_study_4-a}, we can find that the reason why one-shot effect is worse than zero-shot effect is that the example we give always have inevitable biases, which will induce the VLMs to make wrong judgments, even worse than the VLMs' own judgment under zero-shot setting. Furthermore, VLMs have serious hallucination problems so in one-shot experiments settings, there is always a tendency to output example answers as part of the results. 

Second, by comparing predictions in Figure.~\ref{fig:case_study_4-b}, we can find that the \approach makes an error to directly output the address of the table range, while \approach correctly output the values on the four boundaries of the table. According to the previous experiment results, we learn that the VLMs has poor spatial perception of spreadsheet images, so it's hard for them to infer the address of table ranges accurately. In contrast, VLMs has quite strong OCR capabilities, which allow to decode the cell values on the table boundaries.

\end{document}